\pdfoutput=1

\documentclass[11pt]{article}

\usepackage[preprint]{acl}

\usepackage{times}
\usepackage{latexsym}

\usepackage[T1]{fontenc}

\usepackage[utf8]{inputenc}

\usepackage{microtype}

\usepackage{inconsolata}

\usepackage{graphicx}
\usepackage{color}
\usepackage{latexsym}
\usepackage{makecell}
\usepackage{amsmath,amssymb,bm}
\usepackage{url}
\usepackage {diagbox}
\usepackage[ruled, linesnumbered, titlenumbered]{algorithm2e}
\usepackage{bbm}
\usepackage{multirow}
\usepackage {booktabs}
\usepackage{pifont}
\usepackage{siunitx}
\usepackage{array}
\usepackage{colortbl}
\usepackage[utf8]{inputenc}
\usepackage{CJKutf8}
\usepackage{type1ec}
\usepackage[T2A,T1]{fontenc}
\usepackage[russian, portuges, ngerman, french, dutch, english]{babel}
\usepackage{textcomp}
\DeclareRobustCommand{\textcyr}[1]{\foreignlanguage{russian}{#1}}

\usepackage{caption}
\usepackage{footnote}
\makesavenoteenv{tabular}

\newcommand*{\ja}[1]{\begin{CJK}{UTF8}{ipxm}#1\end{CJK}}
\newcommand*{\zh}[1]{\begin{CJK}{UTF8}{gbsn}#1\end{CJK}}

\title{mCSQA: Multilingual Commonsense Reasoning Dataset \\with Unified Creation Strategy by Language Models and Humans}

\author{Yusuke Sakai, Hidetaka Kamigaito, Taro Watanabe \\
  Nara Institute of Science and Technology  \\
  \texttt{\{sakai.yusuke.sr9, kamigaito.h, taro\}@is.naist.jp}}

\begin{document}
\maketitle
\begin{abstract}

It is very challenging to curate a dataset for language-specific knowledge and common sense in order to evaluate natural language understanding capabilities of language models. Due to the limitation in the availability of annotators, most current multilingual datasets are created through translation, which cannot evaluate such language-specific aspects.
Therefore, we propose Multilingual CommonsenseQA (mCSQA) based on the construction process of CSQA but leveraging language models for a more efficient construction, e.g., by asking LM to generate questions/answers, refine answers and verify QAs followed by reduced human efforts for verification. 
Constructed dataset is a benchmark for cross-lingual language-transfer capabilities of multilingual LMs, and experimental results showed high language-transfer capabilities for questions that LMs could easily solve, but lower transfer capabilities for questions requiring deep knowledge or commonsense. 
This highlights the necessity of language-specific datasets for evaluation and training. Finally, our method demonstrated that multilingual LMs could create QA including language-specific knowledge, significantly reducing the dataset creation cost compared to manual creation.
The datasets are available at \url{https://huggingface.co/datasets/yusuke1997/mCSQA}.

\end{abstract}

\section{Introduction}

\begin{figure}[t]
\centering
\includegraphics[width=0.947\linewidth]{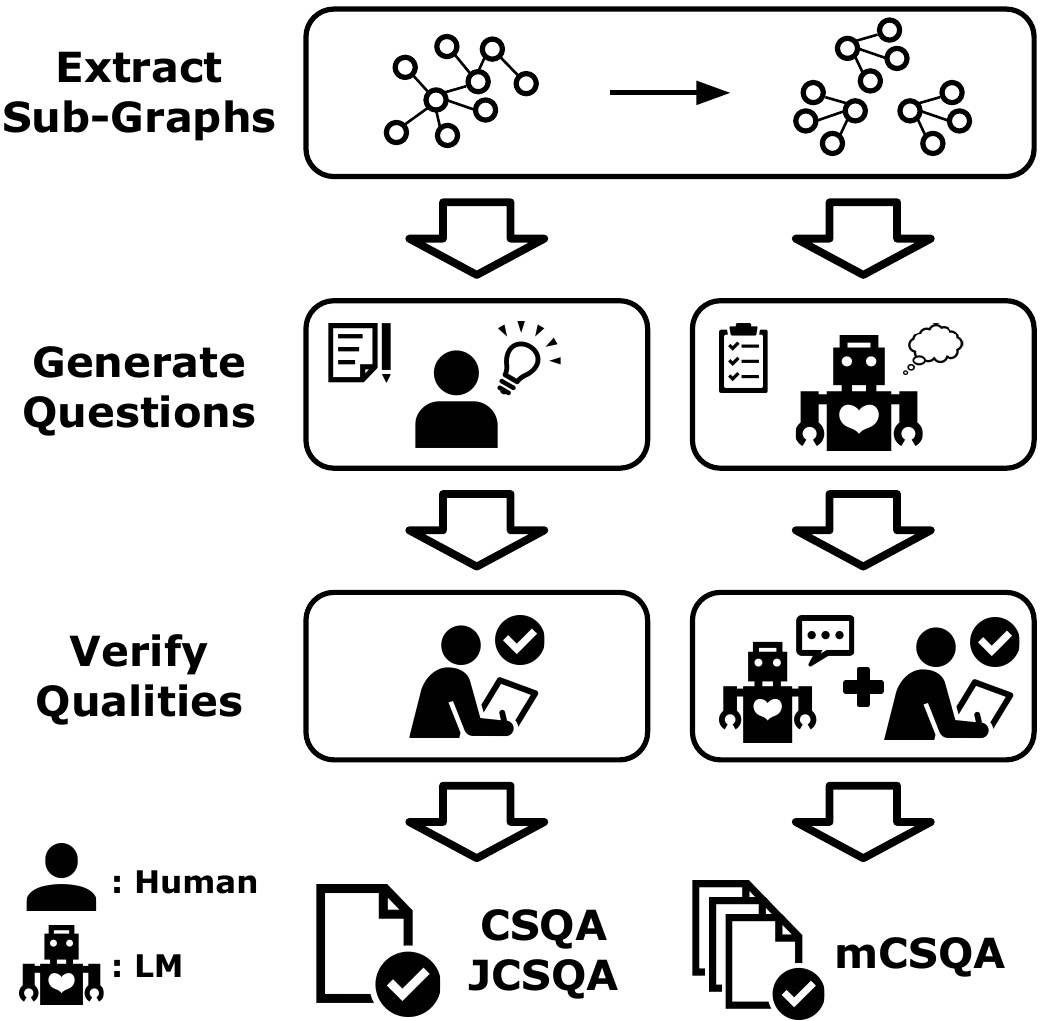}
\caption{The comparison of the dataset creation process for mCSQA and (J) CSQA includes two key changes for efficient and low-cost creation of multilingual datasets. First, the question generation process shifts from human annotators to an LM. Second, an LM assists humans for the quality verification process.}
\label{fig:main-sub}
\end{figure}

\begin{table}[!t]
\centering
\small
\setlength{\tabcolsep}{3pt}
\begin{tabular}{@{}lllll@{}}
\toprule
\multicolumn{5}{c}{Japanese Question (Translated to English)}\\
\midrule
\multicolumn{5}{l}{Q: \ja{お年寄りは？}} \\
\multicolumn{5}{l}{(Who is the elderly person?)} \\
(a) \makecell[tl]{\ja{わし}\\(me)} &(b) \makecell[tl]{\ja{わたし} \\(me)}&(c) \makecell[tl]{\ja{ぼく} \\(me)}& (d) \makecell[tl]{\ja{おれ}\\(me)} &(e)  \makecell[tl]{\ja{うち} \\(me)}\\
\midrule
\multicolumn{5}{c}{English Question (Translated to Japanese)}\\
\midrule
\multicolumn{5}{l}{Q: How do we make a cake?}\\
\multicolumn{5}{l}{\ja{(ケーキを作るにはどうする？)}}\\
(a) \makecell[tl]{roast \\ \ja{(焼く)} \\ (burn)}& (b) \makecell[tl]{broil \\ \ja{(焼く)}\\ (burn)} & (c) \makecell[tl]{grill\\ \ja{(焼く)}\\ (burn)}&  (d) \makecell[tl]{toast \\ \ja{(焼く)}\\ (burn)} & (e) \makecell[tl]{bake\\ \ja{(焼く)}\\ (burn)}\\
\bottomrule
\end{tabular}
\caption{Examples require language-specific knowledge. They cannot be solved without such knowledge, as the translations consolidate the nuances into a single word.}
\label{tab:example-ambi}
\end{table}


Can you choose the correct answer in Table~\ref{tab:example-ambi}? Each choice is semantically very close, making it difficult for non-native speakers to distinguish them. However, native speakers who have language-specific commonsense and knowledge can choose the most plausible choice considering subtle nuances. Despite the need to consider different backgrounds for each language, the datasets to evaluate the natural language understanding (NLU) capabilities of language models (LMs) are mostly for a few major languages such as English, and thus, many languages lack such datasets. When focusing on the cross-lingual capability of LMs, datasets created from scratch in multiple languages are limited, and currently, evaluations mostly use datasets created through translation. However, as can be seen from the example in Table~\ref{tab:example-ambi}, datasets created through translation cannot accurately evaluate language-specific commonsense or knowledge. Therefore, it is necessary to create datasets for each language from scratch, but the manual creation of such datasets is limited by the availability of annotators and financial costs.

To tackle this problem, as shown in Figure~\ref{fig:main-sub}, we propose a method to efficiently create multilingual NLU datasets from multilingual resources by replacing some of the manual annotation processes with generative multilingual LMs. In this study, we focus on CommonsenseQA (CSQA)~\cite{talmor-etal-2019-commonsenseqa}, a dataset for evaluating commonsense reasoning capabilities within NLU evaluations. CSQA is a major commonsense reasoning Question-Answering dataset manually created from the multilingual knowledge base ConceptNet~\cite{conceptnet}. However, due to such limitations, CSQA has been created from scratch only in English and Japanese, JCommonsenseQA (JCSQA)~\cite{kurihara-etal-2022-jglue}. Therefore, we create a Multilingual CommonsenseQA (mCSQA) that extends CSQA to eight languages\footnote{English (en), Japanese (ja), Chinese (zh), German (de), Portuguese (pt), Dutch (nl), French (fr), Russian (ru)} using our proposed method. 

Furthermore, we evaluated the cross-lingual language-transfer capabilities of multilingual LMs focusing on language-specific common sense and knowledge using mCSQA. The results showed high language-transfer capabilities for questions that LMs could easily solve, but lower transfer capabilities for questions requiring deep knowledge or commonsense. 
The total cost per question in mCSQA was reduced to one-hundredth of that for CSQA.

To summarize, our contributions are as follows:
\begin{itemize}
    \item We propose an efficient and low-cost method for creating NLU datasets by generative multilingual LMs.
    \item We demonstrate the potential effectiveness of using multilingual LMs for creating datasets from multilingual resources.
    \item mCSQA makes it possible to analyze the cross-linguistic commonsense understanding capabilities and transfer performance from each language beyond English. 
    \item The analysis revealed that, when focusing on language transfer capabilities using mCSQA, we identified cases where language-specific knowledge is required and cases where it is not, thereby confirming the need for non-translated language-specific datasets.
\end{itemize}

\section{Background and Related Work}
\label{sec:background-related-work}
\paragraph{Commonsense reasoning task}
This task evaluates how an LM can understand and infer object recognition, visual information, and cultural or societal common sense, which are not typically described in textual information. CSQA is a multiple-choice question task that asks for the most plausible choice as an answer with some variants: JCSQA is in Japanese, CommonsenseQA 2.0~\cite{talmor2021commonsenseqa} is a more challenging dataset, ECQA~\cite{aggarwal-etal-2021-explanations} requires explaining the process of deriving an answer, etc. There exist other types of commonsense tasks: COPA~\cite{roemmele2011choice} and BalancedCOPA~\cite{kavumba-etal-2019-choosing} ask about causal relationships between everyday events; SocialIQA~\cite{sap-etal-2019-social} asks about social common sense; PIQA~\cite{bisk2020piqa} evaluates procedural knowledge; HotpotQA~\cite{yang-etal-2018-hotpotqa} requires multi-hop inference; DROP~\cite{dua-etal-2019-drop} captures arithmetic operation capabilities; and tasks like understanding language information~\cite{liu-etal-2022-testing, kocijan2023defeat, sakaguchi2021winogrande, wang2018glue}, understanding causal relationships within documents~\cite{mostafazadeh-etal-2020-glucose, mostafazadeh-etal-2016-corpus, zhang2018record, huang-etal-2019-cosmos, ostermann-etal-2018-mcscript, smirnov-2019-neural}, and CommonGen~\cite{lin-etal-2020-commongen}, which asks to generate common sentences from given keywords. The above datasets primarily focus on English, but there exist datasets in Japanese~\cite{omura-etal-2020-method, takahashi-etal-2019-machine, hayashibe-2020-japanese}, Chinese~\cite{xu-etal-2021-blow, xu-etal-2020-clue, wang-etal-2022-cn}, Russian~\cite{shavrina-etal-2020-russiansuperglue, taktasheva-etal-2022-tape}, and Indonesian~\cite{koto-etal-2022-cloze}. For multilingual datasets, most are extended versions of existing ones through translation, such as X-COPA~\cite{ponti-etal-2020-xcopa} from COPA, X-CSQA~\cite{lin-etal-2021-common} from CSQA, and X-CODAH~\cite{lin-etal-2021-common} from CODAH~\cite{chen-etal-2019-codah}. A few datasets, such as TyDiQA~\cite{clark-etal-2020-tydi}, are created for each language from scratch.

\paragraph{Multilingual datasets}
\begin{table}[t]
\centering
\small
\setlength{\tabcolsep}{3pt}
\begin{tabular}{@{}lccc@{}}
\toprule
Methods & Knowledge& Alignment & Costs\\
\midrule
By translation    & \textcolor{red}{\ding{55}} & \textcolor{green}{\ding{51}} & \textcolor{green}{\ding{51}} \\
Compilation of similar tasks   & \textcolor{green}{\ding{51}} & \textcolor{red}{\ding{55}} & \textcolor{green}{\ding{51}} \\
From multilingual resources    &\textcolor{green}{\ding{51}} & \textcolor{green}{\ding{51}} & \textcolor{red}{\ding{55}} \\
\midrule
Ours  &\textcolor{green}{\ding{51}} & \textcolor{green}{\ding{51}} & \textcolor{green}{\ding{51}} \\
\bottomrule
\end{tabular}
\caption{Categorize the multilingual datasets creation methods.}
\label{tab:compare}
\end{table}
When focusing on the evaluation of multilingual performance of LMs, the evaluation datasets are almost exclusively created through three methods, as shown in Table~\ref{tab:compare}: (1) Translation from existing datasets in a major language, e.g., English~\cite{lin-etal-2021-common, ponti-etal-2020-xcopa, conneau-etal-2018-xnli, artetxe-etal-2020-cross, yang-etal-2019-paws}; (2) Compilation of similar tasks across multiple languages~\cite{zhang2023mela, hu2023revisiting, adelani-etal-2022-masakhaner, roy-etal-2020-lareqa, malmasi-dras-2015-large}; (3) Creation from multilingual resources following the same dataset creation process~\cite{keung-etal-2020-multilingual, huang-etal-2020-multilingual, buchholz-marsi-2006-conll, clark-etal-2020-tydi, schwenk-li-2018-corpus, kabra-etal-2023-multi}.
However, (1) translated datasets often do not account for language-specific culture, knowledge, common sense, or linguistic phenomena, leading to a bias towards the background of the source language~\cite{hu-etal-2021-investigating, lin-etal-2021-common, acharya2020atlas, clark-etal-2020-tydi, park2021klue, kurihara-etal-2022-jglue}. 
(2) Simply compiling datasets curated for each individual language could allow the evaluation of language-specific knowledge and common sense.
However, it is difficult to align tasks across languages since most tasks differ in their creation methods data sources or philosophies.
Thus, it just leads to evaluating the transfer capability among comparable tasks, and not evaluating the true transfer capabilities across languages. 
Therefore, (3) only the datasets created from multilingual resources can enable the evaluation of language transfer capability, considering the differences in language-specific knowledge and common sense. Nevertheless, the manual creation of such datasets is limited by the availability of annotators and financial costs.

\paragraph{Dataset creation with LMs}
The superior performance of generative language models allows to create datasets automatically.
SWAG~\cite{zellers-etal-2018-swag} and HellaSwag~\cite{zellers-etal-2019-hellaswag} have created answer choice options through the output of LMs. Such efforts have also been extended to use LMs for data augmentation~\cite{staliunaite2021improving, kumar-etal-2019-closer, kumar-etal-2020-data, lee2021neural}. WANLI~\cite{liu-etal-2022-wanli}, created from MNLI~\cite{williams-etal-2018-broad}, employs GPT-3~\cite{brown2020language} for adversarial data augmentation with manual checks to create challenging datasets. 
Some studies propose methods to manually check quality of LM generation results~\cite{tekiroglu-etal-2020-generating, yuan2021synthbio, wiegreffe-etal-2022-reframing, wang-etal-2021-want-reduce, li-etal-2023-coannotating}. Additionally, there are attempts to create datasets from scratch with emergent abilities of LMs, without using any examples~\cite{he-etal-2022-generate, wang2021zerolabel, schick-schutze-2021-generating, meng2022generating, ye-etal-2022-zerogen}. 
However, these studies have primarily focused on a single language, e.g., English.
Recently, the outputs of language models themselves have been used to create datasets~\cite{honovich-etal-2023-unnatural, zhihong2023synthetic, sun2023principledriven, peng2023instruction} for instruction-tuning~\cite{wei2022finetuned}. TarGEN~\cite{gupta2023targen} employs a single language model and splits the data generation process into multiple steps, inputting the suitable prompt for each step to ensure data diversity and reliability.
\citet{putri2024llm} focus on middle-resource (Indonesian) and low-resource (Sundanese) languages, and investigate whether LLMs can create culturally aware commonsense questions by comparing translation datasets and those generated by LLMs from scratch.

\section{Datasets Creation}
\begin{figure*}[t]
\centering
\includegraphics[width=\linewidth]{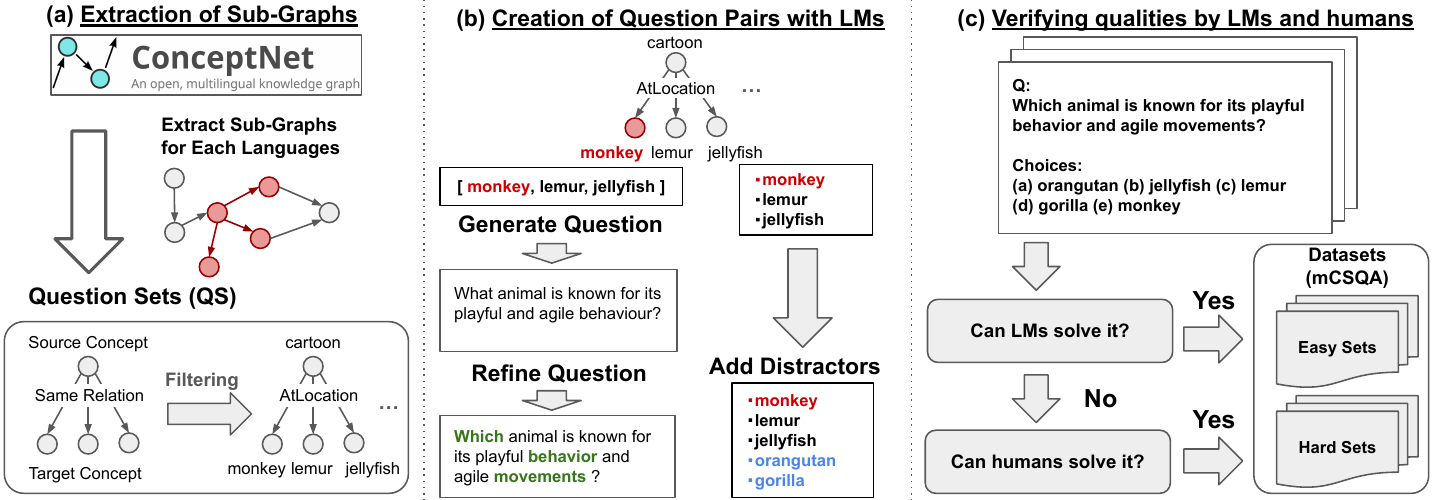}
\caption{Creation process of mCSQA}
\label{fig:main}
\end{figure*}

Our mCSQA construction involves three main steps (see Figure~\ref{fig:main}): extraction of sub-graphs from ConceptNet, creation of question and choice pairs with LMs, and verification of their quality by both LMs and humans. We basically follow the creation processes of CSQA and JCSQA, but modified to allow for unified processing to support multiple languages.

\subsection{Extract Sub-Graphs from ConceptNet}
ConceptNet is a graph knowledge base defined as a tuple, $\mathcal{G}=(\mathcal{C}, \mathcal{R}, \mathcal{T})$, where $\mathcal{C}$ denotes a set of concept entities, $\mathcal{R}$ denotes a set of relations and $\mathcal{T}$ denotes a set of triples. Each triple is represented as $(s, r, t) \in \mathcal{T}$, where $s$ and $t \in \mathcal{C}$ are the source and target concept entities, respectively, and $r \in \mathcal{R}$ is the relation, and carry commonsense knowledge such as ``(student, CapableOf, forget to do homework)''.

We extract subgraphs from ConceptNet, as per Figure~\ref{fig:main}-(a), that have three distinct concept entities derived from queries of concept entities and relations for each language. CSQA uses only forward queries $(s, r, ?)$, but, similar to JCSQA, we also utilize backward queries $(?, r, t)$. We name this subgraph as Question Sets (QSs). After extraction, we filter the QSs like CSQA and JCSQA, and applies unified filtering in mCSQA as follows:

\begin{enumerate}
    \item Similar to CSQA, we retain only QSs that contain any types of the 22 relations\footnote{Antonym, AtLocation, CapableOf, Causes, CausesDesire, DefinedAs, DerivedFrom, Desires, DistinctFrom, EtymologicallyDerivedFrom, HasA, HasFirstSubevent, HasLastSubevent, HasPrerequisite, HasProperty, InstanceOf, MadeOf, MotivatedByGoal, NotDesires, PartOf, SymbolOf, UsedFor}.
    \item We filter out QSs where any of the concept entities consist of more than four words or only a single character\footnote{Unsegmented languages, like Japanese, are segmented by morphology in ConceptNet, so we can apply similar filtering.}.
    \item We remove QSs where any pair of concept entities is connected by a `Synonym' relation in ConceptNet, or where entities are substrings of each other.
\end{enumerate}
After filtering with the above settings, we randomly selected 6,000 QSs for each language\footnote{For French and Russian, the number of QSs did not reach 6,000, so we used all available QSs, totaling 4,125 and 3,901, respectively.}.

\subsection{Create Questions with LMs}
\label{sec:create_questions_with_LMs}

\begin{table}[t]
\centering
\small
\setlength{\tabcolsep}{3pt}
\begin{tabular}{@{}lccc@{}}
\toprule
step& temperature& top\_p & seed\\
\midrule
Creating question sentences     & 0.0 & 0.0 & 0 \\
Refining question sentences     & 0.7 & 0.5 & 0 \\
Adding additional distractors  & 1.2 & 0.7 & 0 \\
\bottomrule
\end{tabular}
\caption{The hyper-parameters for each step}
\label{tab:openai-parameter}
\end{table}

We employ the generative multilingual language model GPT-3.5\footnote{We used gpt-3.5-turbo-1106.}~\cite{NEURIPS2022_b1efde53_gpt3.5} to generate questions automatically to eliminate the human labor as done in CSQA and JCSQA.

Our construction process comprises three steps of `question generation', `question refinement' and `distractor augmentation' as shown in Figure~\ref{fig:main}-(b). Our step differs from CSQA in the refinement step since we need to improve the question generation from LM.

We designed prompts and tuned optimized hyper-parameters for each step for LMs. The details of the prompts are described in Appendix~\ref{sec:detail-of-prompts}, and the hyper-parameters are shown in Table~\ref{tab:openai-parameter}.

\paragraph{Creating question sentences}
For each QS, we generated question sentences by LMs where, for each of the three target concept entities, only one serves as the answer. The prompt for LMs was inspired by the JCSQA filtering process for question creation in which systematic filtering uses textual information. The key instructions are as follows:

\begin{itemize}
    \item Avoid including words of the target entities in the question sentence.
    \item Avoid using superficial information such as character count.
    \item End the sentence with a question mark (?).
    \item Be an objective question sentence.
    \item Consists of only one sentence.
\end{itemize}

After generating questions with LMs, we removed any question sentences that do not follow these instructions or contain inappropriate expressions through pattern matching\footnote{We detected inappropriate expressions using \url{https://platform.openai.com/docs/guides/moderation}.}.

\paragraph{Refining question sentences}

LMs do not always generate appropriate outputs resulting in unnatural expressions or degeneration~\cite{Liu2022KGR4, honovich-etal-2023-unnatural, raunak-etal-2023-leveraging, lin-etal-2020-commongen, madaan2023selfrefine}. Hence, inspired by the idea of output refinement~\cite{Liu2022KGR4, raunak-etal-2023-leveraging,  madaan2023selfrefine}, we refine unnatural generated question sentences into natural ones using the LM again and remove inappropriate questions as done in the previous step. Table~\ref{tab:refine-percentage} shows the percentage of sentence refinement.

\begin{table}[t]
\centering
\resizebox{\linewidth}{!}{%
\small
\setlength{\tabcolsep}{1.2pt}
\begin{tabular}{@{}lrrrrrrrr@{}}
\toprule
& en& ja & zh & de & pt & nl & fr & ru\\
\midrule
Total    & 14,722 & 15,695 & 17,254 & 16,542 & 16,679 & 15,992 & 10,770 & 10,215\\
Refined  & 3,654  & 12,007 & 6,534 & 765 & 585 & 7,927 & 3,109 & 6,734\\
\midrule
pct. (\%)        & 24.82 & 76.50 & 37.87 & 4.63 & 3.51 & 49.57 & 28.87 & 65.92 \\
\bottomrule
\end{tabular}
}
\caption{The percentage of sentences refined}
\label{tab:refine-percentage}
\end{table}

\paragraph{Adding additional distractors}
We added additional incorrect choices to make the task more difficult as done in CSQA and JCSQA, but we leverage LM, not crowd workers, to formulate distractors that seemed plausible or related to the questions.
Here, we asked LM to generate two plausible distractors given the three choices of a question without question itself in order to separate the question generation and answering capabilities of LMs.
There is a risk of generating duplicated choices or adding correct choices since question sentence itself is not fed in this process.
Hence, we remove such questions through manual verification in Section~\ref{sec:filtering}.

\subsection{Question Quality Verification by LMs and Humans}
\label{sec:filtering}

In CSQA and JCSQA, every question is manually verified to remove low-quality questions, such as those with multiple correct answers or without correct answers in the choices. However, due to the large number of questions, manually verifying every question is not practical. 
Thus, we leverage simple active learning methodologies for annotation~\cite{liu-etal-2022-wanli, bartolo-etal-2022-models, li-etal-2023-coannotating, kratzwald-etal-2020-learning}. As shown in Figure~\ref{fig:main}-(c), initially, the LM verifies whether the questions can be answered or not, and only those questions that the LM cannot answer are manually verified.

\paragraph{Verification by LMs}
The original questions can be categorized into three types: those questions 1) which are correctly answerable by LMs, 2) which are wrongly answered by LMs, but humans can choose the correct one, 3) which are not answerable either by LMs or humans due to flaws in the question.
Therefore, first, we identify the set of questions LMs can answer, and then manually verify the questions that LMs could not answer correctly to remove flawed questions.
\paragraph{Verification by Humans}
 We hired two crowd workers per language via Amazon Mechanical Turk (MTurk)\footnote{There are workers for each language on MTurk~\cite{pavlick-etal-2014-language}. We hired workers who have an approval rate greater than 90\% with at least 50 approved HITs.}. The crowd workers were presented with the question sentence, choices, and answer, and they were asked to verify if the answer could be concluded from the question and choices. We retained only those questions on which all crowd workers agreed.

\subsection{Data Splitting and Statistics}
\begin{table}[t]
\centering
\resizebox{\linewidth}{!}{%
\small
\setlength{\tabcolsep}{3pt}
\begin{tabular}{@{}lrrrrrrr@{}}
\toprule
\multicolumn{1}{c}{} & \multicolumn{1}{c}{Train} & \multicolumn{3}{c}{Dev} & \multicolumn{3}{c}{Test}\\
\cmidrule(l){2-2} \cmidrule(l){3-5} \cmidrule(l){6-8}
&Total& Easy & Hard & Total & Easy & Hard & Total\\
\midrule
English    & 10,910 & 1,071 & 292 & 1,363 & 1,071 & 292 & 1,363 \\
Japanese   & 11,696 & 1,117 & 344 & 1,461 & 1,117 & 344 & 1,461 \\
Chinese    & 12,159 &   972 & 546 & 1,518 &   972 & 546 & 1,518 \\
German     & 12,504 & 1,279 & 283 & 1,562 & 1,279 & 283 & 1,562 \\
Portuguese & 12,659 & 1,234 & 348 & 1,582 & 1,234 & 348 & 1,582 \\
Dutch      & 12,215 & 1,255 & 271 & 1,526 & 1,255 & 271 & 1,526 \\
French     &  8,047 &   786 & 219 & 1,005 &   786 & 219 & 1,005 \\
Russian    &  6,623 &   445 & 382 &   827 &   445 & 382 &   827 \\
\bottomrule
\end{tabular}
}
\caption{The statistics of mCSQA}
\label{tab:datasets}
\end{table}

Similar to CSQA and JCSQA, we randomly split the data for each language into training, development, and test sets with an 80/10/10 split. 
The mCSQA is evaluated by accuracy following the standard practice in CSQA and JCSQA.
Additionally, in Section~\ref{sec:filtering}, questions that LMs could answer correctly are categorized as Easy, and those answerable by human judgment are categorized as Hard for development and test sets.

\begin{figure}[t]
\centering
\includegraphics[width=\linewidth]{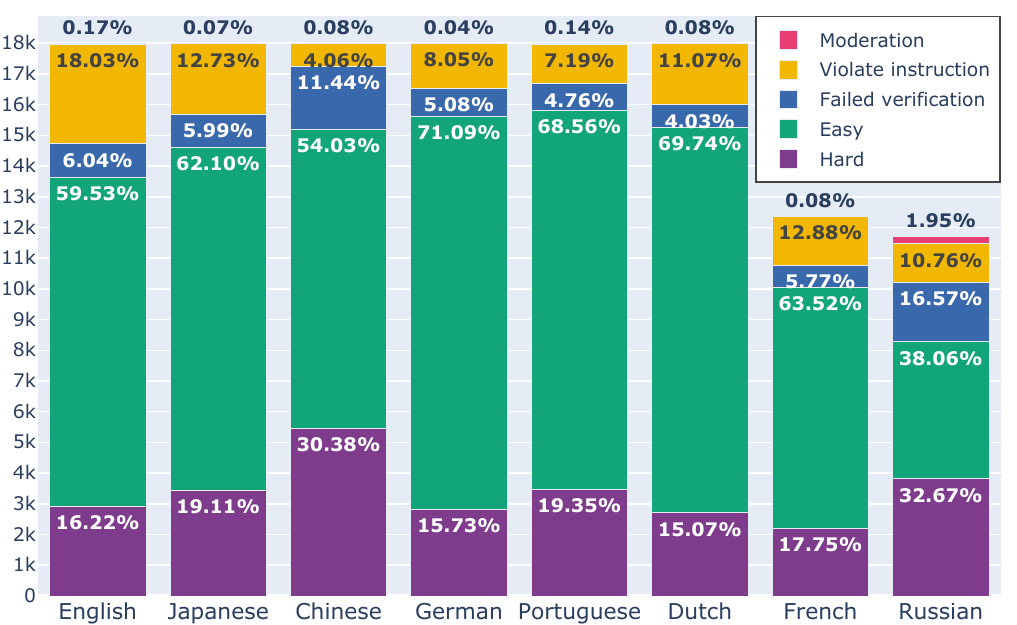}
\caption{The percentage of sentences processed at each step. Easy and Hard were adopted for the dataset, while others were removed during the generation process.}
\label{fig:percentage}
\end{figure}

\begin{figure}[t]
\centering
\includegraphics[width=\linewidth]{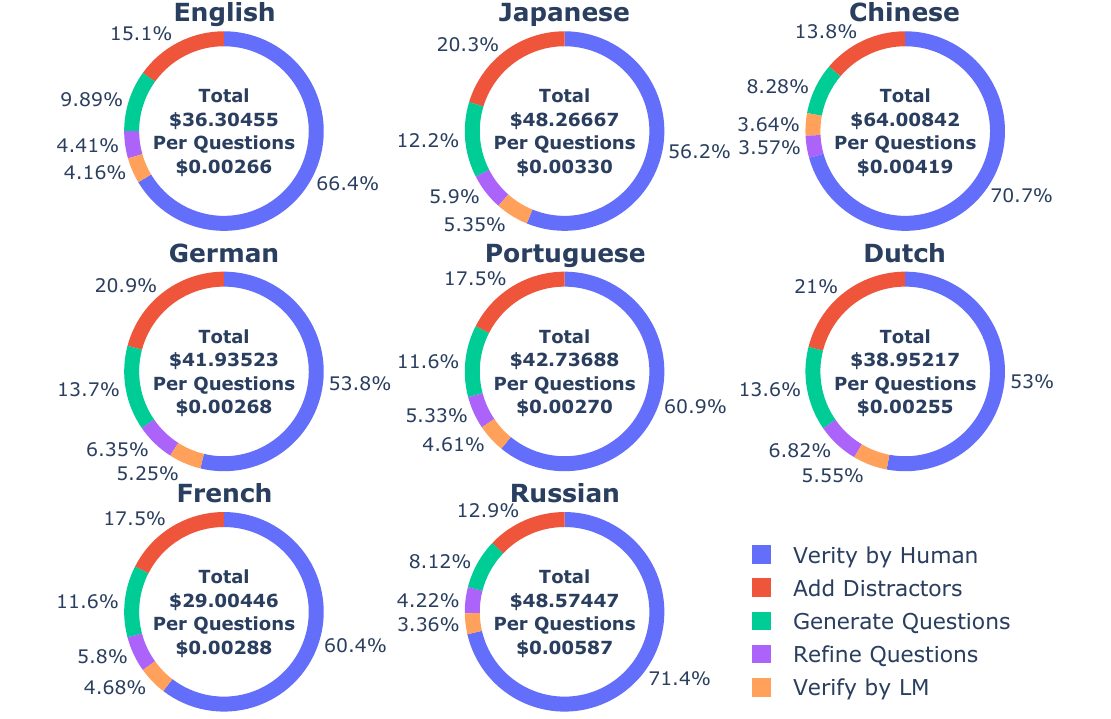}
\caption{The cost details for each language and step.}
\label{fig:fee}
\end{figure}

Table~\ref{tab:datasets} shows the number of questions per language and split, and Figure~\ref{fig:percentage} shows the percentage filtered at each step.
The total cost per question is 0.002 dollars for mCSQA compared to 0.33 dollars for CSQA, reducing the cost to less than one hundredth. Figure~\ref{fig:fee} shows the detailed costs.

Appendix~\ref{sec:mCSQA-quality} discusses more detailed statistics, and Figure~\ref{tab:example-of-mCSQA} shows examples of mCSQA.

\section{Evaluation for mCSQA}
We verify that the mCSQA dataset is meaningful for evaluating the common sense reasoning capability of LMs by using various multilingual LMs.

\subsection{Experimental Setup}

\paragraph{Settings for LMs}

We used mBERT~\cite{devlin-etal-2019-bert}, XLM-100~\cite{NEURIPS2019_c04c19c2_xlm100}, XLM-R~\cite{conneau-etal-2020-unsupervised}, and mDeBERTa-v3~\cite{he2023debertav} as encoder-based multilingual LMs, Llama2-70B~\cite{touvron2023llama}, GPT-3.5~\cite{NEURIPS2022_b1efde53_gpt3.5}, and GPT-4~\cite{openai2023gpt4} as decoder-based multilingual LMs for the experiments. 
Decoder-based LMs inferred with 0-shot and 3-shot settings. For detailed experimental settings, please refer to the Appendix~\ref{sec:detail-of-experimental-settings-appendix}.

\paragraph{Settings for human baseline}
We followed the CSQA setting and randomly selected 100 questions each from the validation and test data for every language to measure the human baseline. We hired five new crowd-workers per language on MTurk. The answers were decided by a majority vote for each question.

\subsection{Evaluation Results}

Table~\ref{tab:result-score} shows the main results.
Focusing on the performance of zero-shot setting of GPT-3.5, which was used for dataset creation, we find that its performance is equivalent to or worse than that of Encoder models like $\text{XLM-R}_{\textnormal{LARGE}}$ and mDeBERTa-v3 except for German and Russian.
When comparing the results of GPT-3.5 with GPT-4, the performance of GPT-3.5 is inferior for most languages to that of GPT-4. This indicates that the questions GPT-3.5 failed to answer correctly are those that cannot be answered by the knowledge of GPT-3.5, and it implies that the root cause is a lack of knowledge or reasoning capability of GPT-3.5.
Furthermore, focusing on Decoder-based models, the results are better in the 3-shot setting than in the 0-shot in most cases. This trend was observed even with the GPT-3.5 used for question creation.

The results show that the prompting technique is effective for mCSQA in exploiting the reasoning capabilities of decoder-based LMs. The trend is similar to other commonsense reasoning tasks like CSQA~\cite{qin-etal-2023-chatgpt, palm, wei2022chain, brown2020language, Dou_Peng_2022}, indicating that mCSQA can be equally effective as a dataset for commonsense reasoning tasks.
Finally, when compared to the human baseline, there is a significant gap in the results of all LMs. Thus, it can be said that even when using LMs for question creation, it is possible to create a dataset with sufficient quality and difficulty for the LMs themselves.

\begin{table*}[h]
\centering
\small
\setlength{\tabcolsep}{3.5pt}
\begin{tabular}{@{}lcccccccccccccccc@{}}
\toprule
  & \multicolumn{2}{c}{English} & \multicolumn{2}{c}{Japanese} &  \multicolumn{2}{c}{Chinese} & \multicolumn{2}{c}{German} & \multicolumn{2}{c}{Portuguese} & \multicolumn{2}{c}{Dutch} & \multicolumn{2}{c}{French} & \multicolumn{2}{c}{Russian}\\ 
  \cmidrule(lr){2-3} \cmidrule(lr){4-5} \cmidrule(lr){6-7} \cmidrule(lr){8-9} \cmidrule(lr){10-11} \cmidrule(lr){12-13} \cmidrule(lr){14-15}  \cmidrule(l){16-17} 
  & dev & test & dev & test & dev & test & dev & test & dev & test & dev & test & dev & test & dev & test \\
\midrule
Human (Rand. 100 sent.) & 87.0 & 93.0 & 89.0 & 95.0 & 91.0 & 87.0 & 96.0 & 96.0 & 93.0 & 93.0 & 98.0 & 97.0 & 96.0 & 92.0 & 87.0 & 94.0\\
\midrule
mBERT-cased & 60.6 & 61.3 & 66.0 & 63.5 & 65.9 & 63.5 & 58.6 & 57.9 & 65.2 & 61.5 & 54.8 & 57.8 & 46.3 & 47.3 & 32.2 & 31.3 \\
mBERT-uncased & 63.4 & 65.2 & 61.3 & 58.9 & 64.0 & 62.0 & 59.3 & 60.3 & 67.6 & 63.9 & 57.3 & 56.9 & 51.1 & 52.4 & 32.5 & 34.0 \\
XLM-100 & 57.2 & 59.0 & 60.2 & 58.8 & 60.0 & 61.5 & 54.4 & 54.7 & 62.7 & 59.5 & 52.2 & 52.0 & 35.3 & 35.0 & 23.2 & 26.0 \\
XLM-R$_{\textnormal{BASE}}$ & 68.0 & 69.1 & 68.5 & 66.2 & 69.8 & 68.3 & 63.9 & 62.8 & 69.5 & 67.3 & 62.0 & 64.0 & 47.6 & 45.5 & 36.9 & 37.0 \\
XLM-R$_{\textnormal{LARGE}}$ & 77.2 & 77.5 & 75.7 & 72.6 & \textbf{75.0} & \textbf{74.1} & 76.2 & 75.4 & 79.0 & 76.4 & 73.0 & 74.7 & 62.0 & 62.3 & 48.9 & 49.5 \\
mDeBERTa-v3 & 76.6 & 79.2 & 77.2 & 74.1 & 74.6 & 72.0 & 75.7 & 77.5 & 78.3 & 78.2 & 72.7 & 74.9 & 62.1 & 62.4 & 51.3 & 49.9 \\
\midrule
Llama2-70B (0-shot) & 48.1 & 47.7 & 25.6 & 24.8 & 26.5 & 25.9 & 32.5 & 32.7 & 38.7 & 37.6 & 40.9 & 39.4 & 42.3 & 44.1 & 23.5 & 22.9 \\
Llama2-70B (3-shot) & 57.1 & 55.5 & 47.4 & 46.6 & 33.3 & 30.2 & 63.1 & 62.9 & 65.0 & 63.7 & 60.8 & 62.3 & 57.8 & 56.7 & 30.8 & 32.3 \\
GPT-3.5 (0-shot) & 76.7 & 77.0 & 76.3 & 76.7 & 64.0 & 63.6 & 81.3 & 81.4 & 77.9 & 77.7 & 82.1 & 81.5 & 78.6 & 77.1 & 53.3 & \textbf{53.0} \\
GPT-3.5 (3-shot) & 77.2 & 78.4 & 77.5 & 77.0 & 65.3 & 64.3 & \textbf{83.2} & 81.4 & 78.5 & 78.0 & 81.8 & 80.5 & 78.4 & 76.5 & \textbf{54.1} & 50.1 \\
GPT-4 (0-shot) & \textbf{80.9} & 80.9 & 78.4 & 77.2 & 66.0 & 65.6 & 81.0 & 81.0 & 78.6 & 77.6 & \textbf{83.4} & 81.5 & 78.8 & 77.0 & 49.9 & 47.8 \\
GPT-4 (3-shot) & 80.5 & \textbf{81.0} & \textbf{78.5} & \textbf{77.5} & 67.2 & 66.9 & 82.6 & \textbf{81.6} & \textbf{80.5} & \textbf{78.8} & 83.3 & \textbf{81.6} & \textbf{79.0} & \textbf{77.4} & 50.1 & 48.9 \\
\bottomrule
\end{tabular}
\caption{The results on mCSQA (acc. \%)}
\label{tab:result-score}
\end{table*}

\begin{figure*}[t]
\centering
\includegraphics[width=\linewidth]{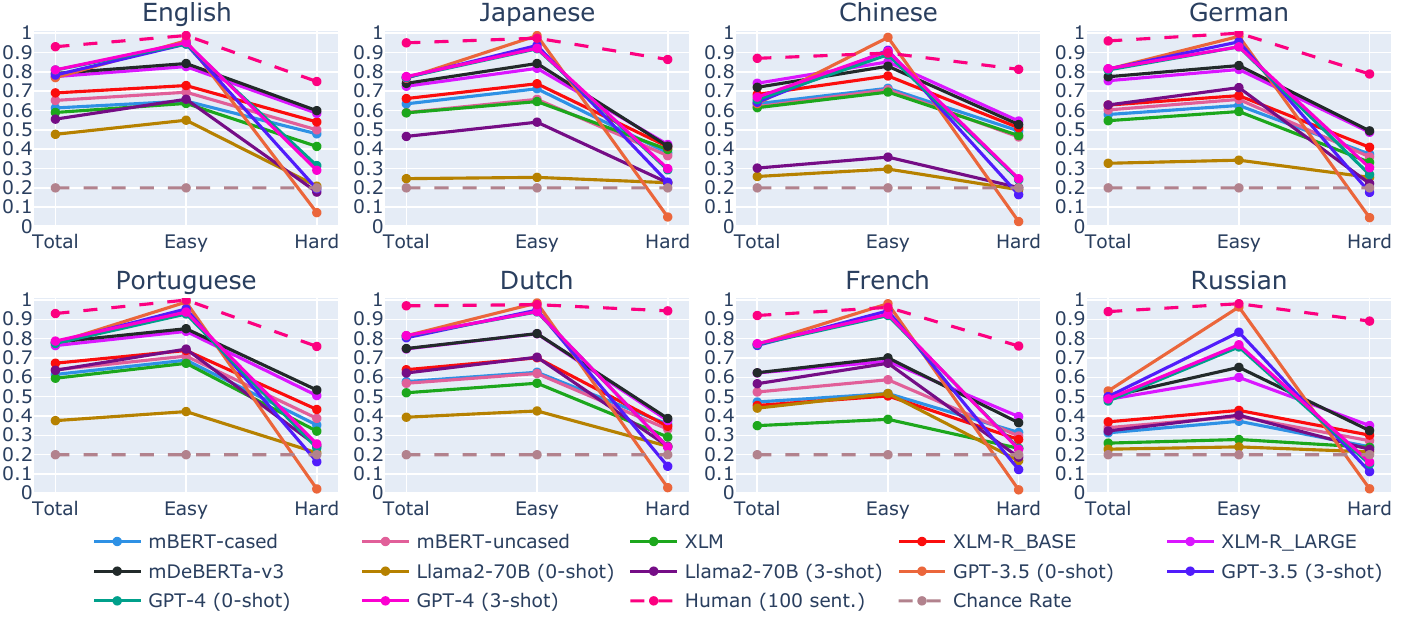}
\caption{Comparison of the evaluation accuracy between Easy and Hard sets.}
\label{fig:easy_hard}
\end{figure*}
\section{Discussion}
\subsection{Comparison of Easy vs. Hard}

We compare the accuracy of Easy and Hard sets for more fine-grained analysis. Figure~\ref{fig:easy_hard} shows the results in the test split. GPT-3.5 and GPT-4 could choose the answer correctly in most cases for the Easy sets, 
but the accuracy is lower in the Hard sets with a significant gap when compared with human results; note that GPT-3.5 cannot answer there sets during the dataset creation.
The other LMs also show a gap in evaluation accuracy with results for Hard sets being lower than those for Easy ones.

These results, specifically the trend observed with GPT-3.5, show that even if LMs can create questions, it does not necessarily mean that they can answer them, and it entails that the question creation and answering are totally different capabilities.
Therefore, we conclude that LMs can substitute for humans in parts of dataset creation processes from structured data and common sense reasoning task creations.

\subsection{Evaluation of Multilingual LMs' Cross-Lingual Transfer Capabilities}

The cross-lingual transfer performance of multilingual LMs is often evaluated from English to other language directions due to linguistic resource reasons. The X-CSQA dataset~\cite{lin-etal-2021-common}, which consists solely of machine-translated questions from CSQA's development and test splits, captures only the one-way cross-lingual transfer performance of LMs that were trained in English to evaluate their performance in other languages. In contrast, mCSQA supports the evaluation of cross-lingual language transfer performance in any directions among multilingual LMs that were trained in each of the eight languages.

\begin{figure}[t]
\centering
\includegraphics[width=\linewidth]{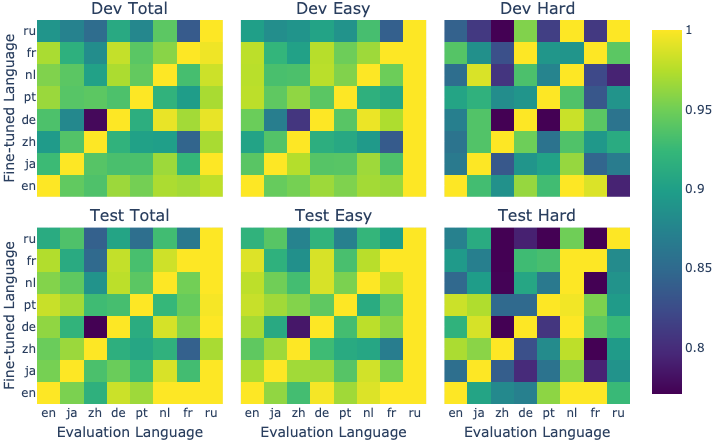}
\caption{The language transfer performance of $\text{XLM-R}_{\textnormal{LARGE}}$. The y-axis indicates the languages in which the model was fine-tuned, while the x-axis indicates the languages used for evaluation. It shows the percentage of performance achieved when compared with the model trained and evaluated in the same language.}
\label{fig:transfer}
\end{figure}

Figure~\ref{fig:transfer} shows the results of the multilingual LM, $\text{XLM-R}_{\textnormal{LARGE}}$, which was fine-tuned in each of the eight languages separately and then evaluated across all eight languages on mCSQA, using the same settings as in Table~\ref{tab:experiment-settings}.
The results from Figure~\ref{fig:transfer} show that, regardless of the language in which they were trained, cross-lingual transfer abilities are observed in most cases for any languages given the relative lower drop of performance when compared with the monolingual performance.
Moreover, in the Easy sets, the drop is within 10\% for most language pairs, while in the Hard sets, it exceeds 20\%. This indicates that questions that are relatively easy to judge (Easy sets) facilitate the language transfer capability, but questions requiring deep background knowledge (Hard sets) necessitate language-specific training and the development of LMs.

\subsection{Which is Better: Monolingual Fine-tuning or Multilingual Fine-tuning?}


\begin{table}[t]
\centering
\footnotesize
\setlength{\tabcolsep}{2.5pt}
\begin{tabular}{@{}ccrrrrrrrr@{}}
\toprule
\multicolumn{1}{c}{} &\multicolumn{1}{c}{} & \multicolumn{8}{c}{Test (\%)}\\
 \cmidrule(l){3-10} 
&      & en & ja & zh & de & pt & nl & fr & ru \\
\midrule
\multirow{4}{*}{Total}& Mono. & 77.5 & 72.6 & 74.1 & 75.4 & 76.4 & 74.7 & 62.3 & 49.5 \\
&Multi. & 81.4 & 74.6 & 74.2 & 77.8 & 79.9 & 77.0 & 65.7 & 54.2 \\
&$\Delta$ & 
\cellcolor[rgb]{1.0, 0.439, 0.439}3.9 &
\cellcolor[rgb]{1.0, 0.714, 0.714}2.0 &
\cellcolor[rgb]{1.0, 0.984, 0.984}0.1 &
\cellcolor[rgb]{1.0, 0.655, 0.655}2.4 &
\cellcolor[rgb]{1.0, 0.498, 0.498}3.5 &
\cellcolor[rgb]{1.0, 0.671, 0.671}2.3 &
\cellcolor[rgb]{1.0, 0.655, 0.655}2.4 &
\cellcolor[rgb]{1.0, 0.325, 0.325}4.7\\ \cmidrule(l){2-10}
& Unseen & 
\cellcolor[rgb]{0.85, 1.0, 0.85}80.0 &
\cellcolor[rgb]{0.85, 1.0, 0.85}71.8 &
\cellcolor[rgb]{0.85, 1.0, 0.85}71.3 &
\cellcolor[rgb]{0.85, 1.0, 0.85}76.6 &
\cellcolor[rgb]{0.85, 1.0, 0.85}76.0 &
\cellcolor[rgb]{0.85, 1.0, 0.85}76.6 &
\cellcolor[rgb]{0.85, 1.0, 0.85}64.9 &
\cellcolor[rgb]{0.85, 1.0, 0.85}51.8\\ \midrule
\multirow{4}{*}{Easy}& Mono. & 82.8 & 81.9 & 85.3 & 81.2 & 83.8 & 82.5 & 68.4 & 60.0\\
&Multi. &86.5 & 83.4 & 85.7 & 84.4 & 86.6 & 84.9 & 73.2 & 70.1 \\
&$\Delta$ & 
\cellcolor[rgb]{1.0, 0.471, 0.471}3.7 &
\cellcolor[rgb]{1.0, 0.639, 0.639}2.5 &
\cellcolor[rgb]{1.0, 0.941, 0.941}0.4 &
\cellcolor[rgb]{1.0, 0.541, 0.541}3.2 &
\cellcolor[rgb]{1.0, 0.600, 0.600}2.8 &
\cellcolor[rgb]{1.0, 0.655, 0.655}2.4 &
\cellcolor[rgb]{1.0, 0.314, 0.314}4.8 &
\cellcolor[rgb]{1.0, 0.000, 0.000}10.1\\ \cmidrule(l){2-10}
& Unseen & 
\cellcolor[rgb]{0.85, 1.0, 0.85}85.4 &
\cellcolor[rgb]{0.85, 1.0, 0.85}81.2 &
\cellcolor[rgb]{0.85, 1.0, 0.85}84.1 &
\cellcolor[rgb]{0.85, 1.0, 0.85}83.4 &
\cellcolor[rgb]{0.85, 1.0, 0.85}83.1 &
\cellcolor[rgb]{0.85, 1.0, 0.85}84.1 &
\cellcolor[rgb]{0.85, 1.0, 0.85}71.8 &
\cellcolor[rgb]{0.85, 1.0, 0.85}68.8 \\ \midrule
\multirow{4}{*}{Hard}& Mono. & 58.6 & 42.4 & 54.6 & 49.1 & 50.6 & 37.6 & 39.7 & 35.1\\
&Multi. & 62.7 & 45.9 & 53.7 & 47.7 & 56.0 & 40.2 & 38.8 & 35.6 \\
&$\Delta$ & 
\cellcolor[rgb]{1.0, 0.412, 0.412}4.1 & 
\cellcolor[rgb]{1.0, 0.498, 0.498}3.5 & 
\cellcolor[rgb]{0.871, 0.871, 1.0}-0.9 & 
\cellcolor[rgb]{0.8, 0.8, 1.0}-1.4 & 
\cellcolor[rgb]{1.0, 0.227, 0.227}5.4 & 
\cellcolor[rgb]{1.0, 0.482, 0.482}3.6 &
\cellcolor[rgb]{0.871, 0.871, 1.0}-0.9 &
\cellcolor[rgb]{1.0, 0.925, 0.925}0.5 \\ \cmidrule(l){2-10}
& Unseen & 
\cellcolor[rgb]{0.85, 1.0, 0.85}60.3 &
\cellcolor[rgb]{0.85, 1.0, 0.85}41.0 &
\cellcolor[rgb]{0.85, 1.0, 0.85}48.7 &
\cellcolor[rgb]{0.85, 1.0, 0.85}45.9 &
\cellcolor[rgb]{0.85, 1.0, 0.85}51.2 &
\cellcolor[rgb]{0.85, 1.0, 0.85}39.1 &
\cellcolor[rgb]{0.85, 1.0, 0.85}38.6 &
\cellcolor[rgb]{0.85, 1.0, 0.85}31.7 \\
\bottomrule
\end{tabular}
\caption{The performance comparison of $\text{XLM-R}_{\textnormal{LARGE}}$ on test data for each language when trained on monolingual training data versus multilingual data. $\Delta$ means the differences in performance between the two settings. Unseen means the accuracy when trained on all training data except for the evaluation language.}
\label{tab:mono-vs-multi}
\end{table}

Some studies~\cite{tran-bisazza-2019-zero, dhamecha-etal-2021-role, trotta-etal-2021-monolingual-cross, barbieri-etal-2022-xlm, portelli2023ailab} reported that multilingual fine-tuning could improve a part of NLU task performance more than monolingual tuning alone. On the other hand, several studies~\cite{tsai-etal-2019-small, kondratyuk-2019-cross, ronningstad-2023-uio, kondratyuk-straka-2019-75} reported that it did not always improve performance in some tasks. We analyzed whether multilingual fine-tuning is effective for commonsense reasoning tasks through mCSQA. We used the whole shuffled training split data in all languages and fine-tuned $\text{XLM-R}_{\textnormal{LARGE}}$ with the same setting as in Table~\ref{tab:experiment-settings}. Table~\ref{tab:mono-vs-multi} compares the accuracy between monolingual fine-tuning, where tuning and evaluation are in the same language, and multilingual fine-tuning, where tuning is performed for all languages, evaluated for each language's accuracy score. These results show that most languages observed improvements, especially in all cases in Easy sets. However, in Hard sets, some cases observed a decline in performance compared to the monolingual setting. Therefore, while training in a multilingual setting generally promotes accuracy improvement, multilingual training might lead to the loss of language-specific commonsense information for questions requiring more human commonsense. This analysis complements the previous reports~\cite{dhamecha-etal-2021-role, zhang-etal-2023-zbl2w, hu-etal-2021-investigating, mueller-etal-2020-sources} on the successes and failures of multilingual training.

Furthermore, Table~\ref{tab:mono-vs-multi} shows the evaluation results of cross-lingual performance in the unseen setting, where the model was not trained on the language for evaluation data. While some languages outperform the monolingual setting, overall results indicate that training with target language data consistently yields better outcomes. 
This suggests that target language data acts as the secret sauce for enhancing NLU performance.
Therefore, it suggests that for language-specific deep knowledge and cultural understanding, language-transfer capability alone is insufficient, and training with datasets focused on language-specific knowledge is necessary.

\subsection{Case Study for Improvement through Few-Shot Learning}

\begin{table*}[t]
\centering
\resizebox{\linewidth}{!}{%
\small
\setlength{\tabcolsep}{2pt}
\begin{tabular}{@{}p{0.75\linewidth}ccc@{}}
\toprule
\multicolumn{1}{c}{Question} & \multicolumn{1}{c}{Answer} & \multicolumn{1}{c}{0-shot} & \multicolumn{1}{c}{3-shot}\\
\midrule
Which types of aquatic animals are commonly found in the open sea? & \multirow{2}{*}{e} & \multirow{2}{*}{a}&\multirow{2}{*}{e}\\
(a) marine life, (b) earth, (c) waves, (d) coastline, (e) oceanic fish & & &\\
\midrule
What is the purpose of using hand gestures while driving? &\multirow{3}{*}{b} & \multirow{3}{*}{c}& \multirow{3}{*}{b}\\
(a) determine what caused noise,  (b) giving signal to,  (c) checking for any potential dangers, &&&\\
(d) warning, (e) investigating the source of the noise &&& \\
\bottomrule
\end{tabular}
}
\caption{Examples of GPT-3.5 correctly answering in a 3-shot setting. In the top example, a 0-shot setting would choose ``marine life'', but considering the phrase ``in the open sea'' in the question, the answer should be narrowed down to ``oceanic fish''. On the other hand, in the bottom example, it chooses ``checking for any potential dangers'', but ``hand gestures while driving'' can include broader, non-dangerous signals such as thank-you gestures. Therefore, the broader ``giving signal to'' is correct. In this way, the 3-shot setting tended to allow for appropriately granular answers that matched the intent of the question.}
\label{tab:example-of-3-shot-improvement}
\end{table*}

As shown in Figure~\ref{fig:easy_hard}, GPT-3.5 correctly answers most questions in the Easy setting of mCSQA, but in the Hard setting, it fails to answer most questions in the 0-shot setting. This is because GPT-3.5 is used for quality filtering of mCSQA in Section~\ref{sec:filtering}, making it inherently unable to answer the questions in the Hard setting in the 0-shot setting. However, in the 3-shot setting, it shows improvement for some questions. Table~\ref{tab:example-of-3-shot-improvement} shows examples of questions correctly answered in the 3-shot setting. Both examples in Table~\ref{tab:example-of-3-shot-improvement} are mainly due to the granularity of the answers. The 3-shot setting promotes answers at an appropriate granularity for questions that are difficult to judge due to inclusive relationships. 

In the top example in Table~\ref{tab:example-of-3-shot-improvement}, careful reading of the questions narrows down the answer choices. On the other hand, in the bottom example, considering various common knowledge in daily life helps to choose the most appropriate answer. Similar characteristics were observed for other languages as well. For more details, qualitative analyses of the mCSQA dataset are described in Appendix~\ref{sec:mCSQA-quality}.

\section{Conclusion and Future Directions}

We proposed an efficient and low-cost method for creating NLU datasets from structured data by utilizing generative LMs as an alternative to traditional human annotation, often crowdsourced. Inspired by CSQA and JCSQA, we created the multilingual commonsense reasoning task dataset, mCSQA, using GPT-3.5 from the structured multilingual knowledge base ConceptNet. We demonstrated that mCSQA is useful for evaluating the commonsense reasoning capabilities of LMs. We also analyzed the language-transfer capability beyond English with mCSQA and examined the language-specific learning from two aspects: question difficulty and language information. Moreover, our study has shown that the use of multilingual LMs enables the construction of multilingual datasets. Therefore, our method can significantly reduce human labor and financial costs.

In this study, we used a single multilingual LM, but since each step is independent, it is possible to replace the LM used in each step with another one. Furthermore, each step can be applied modularly to other methods, making it possible to use this method for creating multilingual datasets, such as those expanded through translation and manual refinement~\cite{yanaka-mineshima-2022-compositional, seo-etal-2022-dog}.
We aim to extend this method to other types of commonsense reasoning tasks and NLU tasks, to efficiently create multilingual data and conduct a more comprehensive analysis of transfer capabilities across a broader range of tasks and languages.

We focused on language-specific commonsense, but languages are shared across various regions. For example, English is spoken in the United States, the United Kingdom, India, Australia, and many other regions each of which is geographically distant and diverse in terms of climate, food, and culture. Therefore, it will be necessary to create more detailed commonsense tasks that consider cultural differences rather than just language such as~\citet{kabra-etal-2023-multi, khanuja2024image, kim-etal-2024-click-benchmark, 10.1162/tacl_a_00634, fung2024massively, lee2024exploring, shwartz-2022-good, hovy-yang-2021-importance, yin-etal-2022-geomlama, shi2024culturebank}. Our dataset construction method can be useful in creating various commonsense reasoning datasets that outgrow language limits.

\section{Limitations}

\paragraph{Data Resources}
The number of multilingual resources is significantly smaller than that of monolingual resources. Additionally, quality is not consistent, and there are imbalances in data volume across languages in these multilingual resources. In this study, we used ConceptNET, a multilingual knowledge base, and encountered these issues as well. For example, despite Spanish having a significantly higher entity count, it obtained fewer QSs due to its inability to meet the required conditions because of ConceptNet's sparsity issue, and thus it was excluded from the language selection for mCSQA. We believe these problems can be addressed through the automatic generation of knowledge bases~\cite{aser-kgbase, ijcai2020p554, west-etal-2022-symbolic, ide2023phalm, nguyen-ccsk} and data augmentation techniques for knowledge bases~\cite{malaviya2020commonsense, ju2022commonsense, wu-etal-2023-commonsense, shen-etal-2023-dense}, supported by their pre-trained knowledge \cite{DBLP:journals/corr/abs-2311-09109}.

\paragraph{Dataset Quality}
In this study, we used GPT-3.5 and simple prompts for data creation. Therefore, there is room for improvement in the selection of LMs and the refinement of prompts. In a pilot study, we tried using GPT-4 and recognized that it is more capable of creating datasets. However, due to budgetary constraints, we have used GPT-3.5 in this study. Thus, it may become possible to create higher quality datasets at a lower cost when the API prices decrease or by switching to other strong LMs such as Gemini~\cite{geminiteam2023gemini}, Mixtral~\cite{jiang2024mixtral}, Llama~\cite{touvron2023llama}, phi~\cite{abdin2024phi3} or Qwen~\cite{bai2023qwen}. Additionally, employing prompt strategies that leverage the capabilities of LMs, such as Chain of Thought (CoT)~\cite{wei2022chain}, Tree of Thought (ToT)~\cite{yao2023tree} and ReAct~\cite{yao2023react}, could potentially lead to the production of higher quality datasets.

\paragraph{Verification of dataset quality by humans}
The human baselines decreased the evaluation result under the Hard sets compared to Easy sets in Figure~\ref{fig:easy_hard}. Therefore, there exists a risk that the Hard sets include flawed questions, even after manual quality verification. The JCSQA has pointed out that such low-quality questions are included in CSQA, and we have confirmed that they are similarly present in JCSQA. Thus, it is extremely difficult to completely eliminate such low-quality questions. Comparing the percentage of data removed in quality verification, CSQA is 25\% (3995/16242), and JCSQA is 19\% (2643/13906), whereas for mCSQA, it is 27\% (604/2226) and 23\% (599/2510) respectively, according to Figure~\ref{fig:percentage} and referenced in their respective papers. This indicates that the filtering ratios are almost comparable when compared to those, showing that this is not a problem unique to mCSQA. Therefore, reinforcing quality verification to filter out low-quality questions is a challenge in our future studies. However, as Figure~\ref{fig:fee} shows, since more than half of the costs are already spent on manual quality verification, simply hiring more crowd workers would not be a better choice. Hence, exploring more efficient methods of quality verification as an alternative or to assist crowd workers in the future is necessary.

\paragraph{Human baseline}
The experimental results include human baselines using small sets of samples. However, \citet{tedeschi-etal-2023-whats} argue that human baselines may lack reliability due to factors such as the payment issues for crowd workers and the impact of random samples. Therefore, it should be noted that the human baselines in this study are merely reference values.

\section{Ethical Considerations}

\paragraph{License}
The mCSQA dataset was created entirely from the outputs of GPT-3.5 and is therefore subject to OpenAI's license terms\footnote{\url{https://openai.com/policies/terms-of-use}}. OpenAI assigns to us all rights, title, and interest in and to the output. As a result, we are retaining the ownership rights. There are no restrictions on distributing the datasets, but using OpenAI's model output to develop models that compete with OpenAI is prohibited. However, it's possible that these terms may change, and there may be a need to impose distribution restrictions depending on the terms.

\paragraph{Moderation}
We eliminated potentially harmful questions such as violence, sexual content, and hate speech by screening through OpenAI moderation APIs\footnote{\url{https://platform.openai.com/docs/guides/moderation}}. However, in the commonsense reasoning dataset, it cannot be guaranteed that it does not include questions that contain societal biases as collective knowledge. This issue has also been pointed out in existing datasets such as CSQA, JCSQA, and other commonsense reasoning datasets, and it is challenging to determine what is considered commonsense constitutes bias~\cite{rajani-etal-2019-explain, sap-etal-2020-social, bauer-etal-2023-social, an-etal-2023-sodapop}. If you encounter any harmful questions that contain such biases, please report them.

\paragraph{Translation Tool}
We used DeepL Pro\footnote{\url{https://www.deepl.com/translator}} to translate the example sentence, especially Table~\ref{tab:example-ambi}, to avoid arbitrary translation. The copyright of the translation sentences belongs to us\footnote{\url{https://www.deepl.com/pro-license}}.

\bibliography{anthology,custom}

\appendix


\section{Details of the Experimental Settings}
\label{sec:detail-of-experimental-settings-appendix}

We used mBERT~\cite{devlin-etal-2019-bert}, XLM-100~\cite{NEURIPS2019_c04c19c2_xlm100}, XLM-R~\cite{conneau-etal-2020-unsupervised}, and mDeBERTa-v3~\cite{he2023debertav} as encoder-based multilingual LMs, Llama2-70B~\cite{touvron2023llama}, GPT-3.5~\cite{NEURIPS2022_b1efde53_gpt3.5}, and GPT-4~\cite{openai2023gpt4} as decoder-based multilingual LMs for the experiments. Table~\ref{tab:models} shows the details of the LMs.
Encoder-based LMs were fine-tuned following the settings in Table~\ref{tab:experiment-settings}. Decoder-based LMs inferred with 0-shot and 3-shot settings\footnote{In the 3-shot setting, the examples were selected randomly from the training data and included both Easy and Hard sets.} with a fixed seed value. For GPT-3.5 and GPT-4, top\_p and temperature were set to 0 to achieve as deterministic outputs as possible. For Llama2-70B, output was generated greedy, and outlines~\cite{willard2023efficient-outlines} were used to fix the output format.

\section{Qualitative Analysis of mCSQA}
\label{sec:mCSQA-quality}
Table~\ref{tab:example-of-mCSQA} shows examples of mCSQAs for each language.
The examples in Table~\ref{tab:example-of-mCSQA} are accompanied by English translations using DeepL\footnote{We are using DeepL Pro (\url{https://www.deepl.com/translator}), therefore, the copyright of the translations belongs to us. (\url{https://www.deepl.com/pro-license})} to avoid arbitrary translation.

\subsection{Can Multilingual LMs Take into Account Language-specific Knowledge?}
\label{sec:example-analysis}

\paragraph{Case study}
When we examine some cases in Table~\ref{tab:example-of-mCSQA}, such as the examples from the Dutch Hard set and the Russian Hard set, we find that the English translations contain duplications among the question choices. However, these duplications arise not from differences in tense or conjugation, but from semantic differences unique to each language, which a native speaker, equipped with language-specific knowledge and common sense, could easily distinguish. Furthermore, in the case of the German Easy sets, knowledge of Germany's unique education system is required, which might be challenging for those unfamiliar with it. Yet, for German speakers, it is common knowledge that such education systems, such as the Abitur\footnote{\url{https://en.wikipedia.org/wiki/Abitur}} related to the Gymnasium\footnote{\url{https://en.wikipedia.org/wiki/Gymnasium_(school)}}, making it answerable for those knowledgeable in German. This demonstrates that multilingual LMs are capable of generating questions that include the kind of language-specific knowledge and common sense that a native speaker would possess.

\begin{table}[t]
\centering
\footnotesize
\setlength{\tabcolsep}{2pt}
\begin{tabular}{@{}lll@{}}
\toprule
\multicolumn{1}{c}{Type} & \multicolumn{1}{c}{Model Name} &  \multicolumn{1}{c}{HuggingFace / OpenAI API} \\
\midrule
\multirow{6}{*}{Encoder} & mBERT-cased & bert-base-multilingual-cased \\
                   & mBERT-uncased & bert-base-multilingual-uncased \\
                   & XLM-100 & xlm-mlm-100-1280 \\
                   & XLM-R$_{\textnormal{BASE}}$ & xlm-roberta-base \\
                   & XLM-R$_{\textnormal{LARGE}}$ & xlm-roberta-large \\
                   &  mDeBERTa-v3 & microsoft/mdeberta-v3-base \\
\midrule
\multirow{3}{*}{Decoder} &Llama2-70B & meta-llama/Llama-2-70b-chat-hf \\
 &GPT-3.5 & gpt-3.5-turbo-1106 \\
 &GPT-4 & gpt-4-1106-preview \\
\bottomrule

\end{tabular}
\caption{Details of the LMs for the experiments.}
\label{tab:models}
\end{table}

\begin{table}[t]
\centering
\footnotesize
\begin{tabular}{@{}lc@{}}
\toprule
Hyper-parameter & Value \\
\midrule
Batch Size & 64 \\
Learning Rate & 2e-5, 3e-5, 5e-5 \\
Seed & 42 \\
Early Stopping &  3 \\
Warmup Ratio & 0.1 \\
Max Sequence Length & 128 \\
\bottomrule
\end{tabular}
\caption{The hyper-parameters used in the experiment, and others, were set to default settings. The implementation used Transformers~\cite{wolf-etal-2020-transformers}.}
\label{tab:experiment-settings}
\end{table}

\paragraph{The effectiveness of the CSQA style QA}

When examining the Japanese Hard set in Table 10, all the choices translate into the names of seafood in English, which does not match the context of a female singer mentioned in the question. Japanese native speakers would normally recognize them as seafood names too, making it seem at first glance that there is no correct answer. However, the correct choice, `\ja{あゆ}' (ayu), when pronounced in Japanese, is read as 'ayu'. This pronunciation is widely known across Japan as the nickname for the famous singer `\ja{浜崎あゆみ}' (Ayumi Hamasaki)\footnote{\url{https://en.wikipedia.org/wiki/Ayumi_Hamasaki}}, making it a plausible choice even though it's not strictly correct. It allows for a satisfactory selection by Japanese native speakers with language-specific knowledge, common sense, and cultural awareness, and is not answerable by English translation only. In Japan, nicknames are often derived from abbreviations of their names or can suggest the names of objects. The distractor `Wakame' is known as the name of a character from the long-running, famous anime `Sazae-san'\footnote{\url{https://en.wikipedia.org/wiki/Sazae-san}} but not as a singer, thus serving its purpose as a distractor in this question effectively. Similarly, if there were a choice like `\ja{いくら}' (common meaning: red caviar; pronounced: ikura), the plausibility of choice in this question might have been divided. Recently, `ikura'\footnote{\url{https://en.wikipedia.org/wiki/Lilas_Ikuta}} has become a popular name, associated with a member of `Yoasobi'\footnote{\url{https://en.wikipedia.org/wiki/Yoasobi}}, a popular artist group among young people. Adding such a choice would confuse the choice of the correct answer because both choices are plausible, so it would not serve effectively as a distractor. This case shows that the choices can define the scope of common sense, thus making the question effective in evaluating common sense accurately.

\subsection{The Relationship between Knowledge, Culture, Commonsense, and Social Bias}
\subsubsection{What is the Commonsense?}

As can be seen from Table~\ref{tab:example-of-mCSQA} and the discussions in section~\ref{sec:example-analysis}, language-specific common sense is closely related to knowledge and culture. The ConceptNet used in this study does not limit the scope of common sense and deals with a wide range of common sense, enabling the inclusion of questions from various backgrounds into mCSQA, following the same trend as CSQA and JCSQA. 

Generally, commonsense not based on the specific culture or knowledge of a language is likely to be a common understanding across all languages, making such problems potentially answerable through the language-transfer ability of multilingual LMs. However, as shown in Table~\ref{tab:example-ambi}, the granularity of actions, events, and behaviors differs by language, which can be considered to be influenced by the cultural background of the language area. 

This study focuses on language-specific common sense that cannot be addressed by translations of datasets from other languages, and the culture and knowledge included in them are shared among native speakers. Therefore, answering questions that require language-specific backgrounds necessitates a certain level of knowledge and culture specific to each language. However, content that is too specialized falls outside the scope of common sense, and common sense and backgrounds vary among individuals. Therefore, we emphasize the precision of coverage in the manual question quality verification steps and employ a majority vote baseline to avoid overly relying on specific knowledge or culture. 

In this way, questions were created that have language-specific common sense which is general for native speakers but not too specialized. If there was a need to create questions asking for knowledge specialized in specific fields, other knowledge bases such as ATOMIC~\cite{sap2019atomic}, and CCSK~\cite{nguyen-ccsk} could be used. However, this study focused on multilingual performance, deeming ConceptNet appropriate for mCSQA.


\subsubsection{Is Commonsense Social Bias?}
Since commonsense includes implicit cognition, it may contain social and cultural biases, and some methods for the removal of explicit and implicit social biases have been proposed~\cite{sap-etal-2020-social, field-tsvetkov-2020-unsupervised, huang-etal-2021-uncovering-implicit, lent-sogaard-2021-common, emelin-etal-2021-moral, bauer-etal-2023-social}. 

Social Chemistry 101~\cite{forbes-etal-2020-social}, BBQ~\cite{parrish-etal-2022-bbq}, and SODAPOP~\cite{an-etal-2023-sodapop} have been proposed for identifying biases within models or for bias detection using LMs. However, it remains challenging to address situations where biased thinking may only emerge when considering multiple-choice QA, where bias does not occur in isolation.

The definition of bias and common sense changes over time and varies from society to society, and what is considered common sense can shift to bias~\cite{lee2023survey}. Therefore, regular updates to the commonsense reasoning datasets are necessary. Our method for generating commonsense reasoning task datasets using LMs allows for low-cost update operations, making it possible to adapt to the changing boundaries between common sense and bias over time. However, this does not fundamentally address the inclusion of bias in datasets.
Moreover, such issues require a deep chain of semantic thinking for resolution, making filtering based on textual information inappropriate. Therefore, it is necessary to develop methods to remove potential biases in commonsense reasoning task datasets in future work.

\section{Discoveries about the LMs Capabilities}

\subsection{Can LMs Create Questions including Commonsense?}

\paragraph{Generation capability}
CommonGen~\cite{lin-etal-2020-commongen} is one of the commonsense reasoning datasets that evaluates whether it is possible to create commonsense sentences from a given set of keywords. According to the leaderboard of CommonGen\footnote{\url{https://github.com/allenai/CommonGen-Eval}}, the performance of GPT-3.5 used in our dataset creation demonstrates a capability for generating commonsense sentences comparable to those written by humans. However, there is still room for improvement in aspects such as word order. Therefore, we introduced refinement steps to encourage corrections in word order and other errors. Since language models have high performance in Grammar Error Correction (GEC)~\cite{loem-etal-2023-exploring, sottana-etal-2023-evaluation, fang2023chatgpt, coyne2023analyzing, kaneko-okazaki-2023-reducing, kwon-etal-2023-beyond}, combining sentence generation from keywords with GEC capabilities in a pipeline helps to compensate for the weaknesses of language models. We believe that the quality of mCSQA questions is at least not inferior to those created by crowd-workers. The capability of multilingual LMs to create commonsense sentences from given keywords has also been demonstrated in the Korean CommonGen~\cite{seo-etal-2022-dog}, indicating that it is possible to generate commonsense sentences multilingually.

\paragraph{Ensuring the quality of questions}
In this study, we have created commonsense reasoning dataset questions using keywords extracted from ConceptNet. Therefore, the language-specific knowledge and commonsense for each language are guaranteed by ConceptNet. Moreover, the LM creates questions following the given instructions through its emergent capabilities from each keyword. To enhance the language-specific performance of the multilingual LM for each language, we have created prompts for each language in this study. As can be seen from the discussion in section~\ref{sec:example-analysis} and Table~\ref{tab:example-of-mCSQA}, it has become possible to generate questions that possess language-specific knowledge. One of the reasons for the capability to create questions with language-specific knowledge may be attributed to the training data of the LM. For example, Wikipedia, one of the common training data for LMs,  has each language which contains descriptions of knowledge unique to that language, so by posing questions in each language, it is thought that knowledge specific to each language is invoked, enabling the generation of questions based on the knowledge of each language. However, this is a hypothesis, and further analysis will be necessary for verification in future work.
Moreover, we have added distractors in addition to the keywords used for generating the question, which means that even if a question can be generated, it may not necessarily be answerable. Furthermore, questions that cannot be answered have been removed, thus ensuring the difficulty and answerability of the QA.

\subsection{Multilingual Capabilities}
\label{sec:polyglot-template}
\paragraph{Is polyglot template effective?}
We translated the prompt to use question generation for each language and tuned it to convey the same meaning in each language in Section~\ref{sec:detail-of-prompts} aimed to emergence the language-specific knowledge. However, it is known that current generative LMs have mainly trained on English, which is better performance for queries made in English. However, several studies~\cite{ahn2022i, shi2023language, wei2022emergent, awasthi-etal-2023-bootstrapping, kasai2023evaluating, jin2023better} show enough performance even if multilingual queries. Note that the reported performance focuses on the ability to answer specific tasks on benchmarks and does not evaluate the emergent multilingual ability, especially question generation. Nevertheless, \citet{whitehouse-etal-2023-llm} shows that the text generation capability beyond English. As shown in Table 10, we were able to generate questions containing language-specific knowledge from the given keywords as intended by using prompts translated into each language. We were able to generate questions that require deep reasoning, including cultural backgrounds and language-specific pronunciation information as shown in Section~\ref{sec:example-analysis}. Therefore, we conclude that using prompts tailored for each language is effective.

\paragraph{Is GPT-3.5 Multilingual LM?}

Yes, some studies~\cite{lai-etal-2023-chatgpt, armengol-estape-etal-2022-multilingual, zhang-etal-2023-dont} have indeed examined multilingual performance, and the training data also includes multilingually\footnote{\url{https://github.com/openai/gpt-3/tree/master/dataset_statistics}}. Therefore, the multilingual capabilities of GPT-3.5, GPT-4, and Llama used in our experiment have also been evaluated~\cite{ahuja-etal-2023-mega, schott-etal-2023-polyglot, chen2024orion14b}, leading us to consider these as multilingual LMs.
However, they still rely predominantly on information from Western norms~\cite{cao-etal-2023-assessing, arora-etal-2023-probing, havaldar-etal-2023-multilingual}, making this issue an ongoing challenge to be addressed in the future.

\paragraph{Exhortation to multilingual instruction-tuning dataset.}
Instruction-tuning~\cite{wei2022finetuned, flan-template, chung2022scaling, wang-etal-2023-self-instruct} can enhance the quality of LMs, e.g. ability to follow instructions and NLU performance. However, in Section~\ref{sec:background-related-work}, the current multilingual datasets include those created through translation, which means that instruction-tuning using such data may not lead to the acquisition of data bias or language-specific knowledge. Given these considerations, the multilingual instruction-tuning data~\cite{kew2023turning, singh2024aya} proposed recently often utilize datasets created through translation, leading to the occurrence of the aforementioned issues to a considerable extent. Consequently, the effectiveness of such instruction-tuning may be diminished. For commonsense reasoning tasks in multilingual instruction-tuning datasets, they sometimes use X-CSQA~\cite{lin-etal-2021-common}. However, since it cannot handle language-specific knowledge or commonsense effectively, it is preferable to use data created from scratch, like mCSQA. Currently, due to data resource issues, reliance on translated data is inevitable, but we hope that in the future, it will be replaced by language-specific data.

\subsection{Hard Sets are Truly Hard?}
The Hard sets consist of questions that the LM used for question creation could not answer, thus reflecting the characteristics of that LM. However, despite the influence of specific LM’s character, a performance decline in the Hard sets compared to the Easy sets was observed across all models. Therefore, while the strict division of sets depends on the model, it has become clear that there is a similar trend across LMs as a whole. For this reason, scoring is conducted without distinguishing between Easy and Hard, using a total score for the entire set, which allows for the absorption of differences due to the models.

\subsection{Generation Bias and Annotation Artifacts}

It has been pointed out that datasets created by LMs contain generation bias~\cite{omura-etal-2020-method, zellers-etal-2019-hellaswag, tamborrino-etal-2020-pre}, and those created by crowd-workers include specific patterns (Annotation Artifacts)~\cite{gururangan-etal-2018-annotation, chen-etal-2019-codah, omura-etal-2020-method}. Annotation artifacts, in particular, have been noted in natural language inference tasks such as MNLI~\cite{williams-etal-2018-broad} and SNLI~\cite{bowman-etal-2015-large}, where choices can be easily distinguished by superficial words like ``not''.

However, \citet{tamborrino-etal-2020-pre} show that the impact of Annotation Artifacts is not present in the CSQA task. Similarly, in this study, we have separated question generation ability and answering ability during the question generation process and shuffled the options, so there are no clues included in the dataset. Moreover, we create Hard sets, even if such biased questions existed, the evaluation is conducted without these biases, allowing for an evaluation that removes these biases.

\begin{table*}[t]
\centering
\resizebox{\linewidth}{!}{%
\small
\setlength{\tabcolsep}{2pt}
\begin{tabular}{@{}llp{0.4\linewidth}ccccc@{}}
\toprule
\multicolumn{2}{c}{Lang.}& \multicolumn{1}{c}{Question} & \multicolumn{5}{c}{Choices} \\
&&& \multicolumn{1}{c}{Correct} & \multicolumn{2}{c}{Distractors} & \multicolumn{2}{c}{Additional Distractors}\\ 
\midrule
EN & Easy & If a cat is feeling irritated, what might it do? & scratch if annoy & look out window & fish with paw & chase a toy & nap in the sun \\ 
\cmidrule(l){3-3} \cmidrule(l){4-4} \cmidrule(l){5-5} \cmidrule(l){6-6} \cmidrule(l){7-7} \cmidrule(l){8-8}
 & Hard & Which animal is known for its playful behavior and agile movements? & monkey & jellyfish & lemur & orangutan  & gorilla  \\ 
\midrule
JA  & Easy  & \ja{音を聞き分けるためには何をしますか？}       & \ja{耳を澄ます}        & \ja{学習する}  & \ja{書き取る} & \ja{実践する} & \ja{経験する} \\ 
    &       & (What do you do to listen to the sounds?) & (Listen carefully)   & (Learn)       & (Write)      & (Practice)  & (Experience) \\ 
\cmidrule(l){3-3} \cmidrule(l){4-4} \cmidrule(l){5-5} \cmidrule(l){6-6} \cmidrule(l){7-7} \cmidrule(l){8-8}
    & Hard  & \ja{日本の女性歌手で、自身の楽曲の作詞・作曲も手がける人気アーティストは誰ですか？}                 & \ja{あゆ}   & \ja{どじょう} & \ja{わかめ} & \ja{うなぎ} & \ja{さけ}\\
    &       & (Which popular Japanese female singer also writes lyrics and composes her own songs?) & (Sweetfish) & (Loach)      & (Wakame)  & (Eel)      & (Salmon) \\
\midrule
ZH  & Easy  & \zh{你在考试前应该做什么？}                  & \zh{回家温习}                      & \zh{聊天}   & \zh{作弊} & \zh{健身}   & \zh{看电影} \\ 
    &       & (What should you do before your exam?)   & \makecell[t]{(Go home and \\study)}  & (Chatting) & (Cheat)   & (Work out) & (Watch films) \\
\cmidrule(l){3-3} \cmidrule(l){4-4} \cmidrule(l){5-5} \cmidrule(l){6-6} \cmidrule(l){7-7} \cmidrule(l){8-8}
    & Hard  & \zh{在感情关系中，最令人痛苦的事情是什么？}                               & \zh{被甩}         & \zh{花大钱}                       & \zh{心碎} & \zh{找到真爱} & \zh{实现梦想} \\
    &       & (What's the most excruciating thing about being in a relationship?) & \makecell[t]{(Getting \\dumped)} & \makecell[t]{(Spending a \\lot of money)} &  \makecell[t]{(Getting your \\heart broken)} & (Finding true love) & \makecell[t]{(Realising your \\dreams)} \\
\midrule
DE  & Easy  & Welche Art von weiterführender Schule bereitet Schüler auf das Abitur vor? & gymnasium        & gesamtschule           & fachoberschule               & berufsschule        & realschule\\
    &       & (What type of secondary school prepares students for the Abitur?)          & (grammar school) & \makecell[t]{(comprehensive \\school)} & \makecell[t]{(technical seco-\\ndary school)} & (vocational school) & (secondary school) \\
\cmidrule(l){3-3} \cmidrule(l){4-4} \cmidrule(l){5-5} \cmidrule(l){6-6} \cmidrule(l){7-7} \cmidrule(l){8-8}
    & Hard  &  Was ist die richtige Bezeichnung für das langsame Abwärtsbewegen auf einer schiefen Ebene? & hinabgleiten & hinabfliegen & dahinab & hinabtauchen & hinabschweben\\
    &       & (What is the correct term for moving slowly downwards on an inclined plane?) & (slide down) & (fly down) & (descend) & (dive down) & (float down) \\
\midrule
PT  & Easy  & Como demonstrar afeto a um animal de estimação? & fazer carinho & alegrar a vida & pentelhar & abraçar & dar um presente \\
    &       & (How do you show affection to a pet?) & (cuddle) & (combing) & (brighten up life) & (cuddle) & (give a gift) \\
\cmidrule(l){3-3} \cmidrule(l){4-4} \cmidrule(l){5-5} \cmidrule(l){6-6} \cmidrule(l){7-7} \cmidrule(l){8-8}
    & Hard  & Qual a ação que um coelho pode fazer para se mover rapidamente? & pular & orientando & segurar & esperar & correr\\
    &       & (What action can a rabbit do to move quickly?) & (jump) & (guiding) & (hold) & (wait) & (run)\\
\midrule
NL  & Easy  & Kunt u mij vertellen wat gokken is? & kansspel & gelijkspel & steekspel & vuurspel & wedstrijd\\
    &       & (Can you tell me what gambling is?) & (game of chance) & (draw) & (joust) & (match) & (fire game) \\
\cmidrule(l){3-3} \cmidrule(l){4-4} \cmidrule(l){5-5} \cmidrule(l){6-6} \cmidrule(l){7-7} \cmidrule(l){8-8}
    & Hard  & Kunt u uitleggen wat een veelvoorkomend begrip is dat verwijst naar iets wat algemeen geaccepteerd of verspreid is in een samenleving? & gemeengoed & gemeenschap & gemeenplaats & gezamenlijk & gebruikelijk \\
    &       & (Can you explain what is a common term that refers to something commonly accepted or widespread in a society?) & (common) & (community) & (commonplace) & (common) & (common) \\
\midrule
FR  & Easy  & Quelle unité de temps correspond à une période de vingt-quatre heures ? & jour & décade & siècle & année & mois\\
    &       &(What unit of time corresponds to a twenty-four hour period?) & (day) & (decade) & (century) & (year) & (month) \\
\cmidrule(l){3-3} \cmidrule(l){4-4} \cmidrule(l){5-5} \cmidrule(l){6-6} \cmidrule(l){7-7} \cmidrule(l){8-8}
    & Hard  & Quelle partie du corps utilise-t-on pour saisir des objets de petite taille ? & doigt & annulaire & auriculaire & majeur & index \\
    &       & (What part of the body is used to pick up small objects?) & (finger) & (ring finger) & (little finger) & (middle finger) & (index finger) \\
\midrule
RU  & Easy  & \textcyr{Какое время года обычно связывается с праздниками Нового года и Рождества?} & \textcyr{зима} & \textcyr{весна} & \textcyr{осень} & \textcyr{летний сезон} & \textcyr{лето}\\
    &       &(What time of year is usually associated with the holidays of New Year's Eve and Christmas?) & (winter) & (spring) & (fall) & (summer season) & (summer) \\
\cmidrule(l){3-3} \cmidrule(l){4-4} \cmidrule(l){5-5} \cmidrule(l){6-6} \cmidrule(l){7-7} \cmidrule(l){8-8}
    & Hard  & \textcyr{Какой звук издает довольный кот?} & \textcyr{урчание} & \textcyr{заурчать} & \textcyr{проурчать} & \textcyr{мурлыкать} & \textcyr{громко урчать} \\
    &       & (What sound does a contented cat make?) & (purr) & (rumble) & (purr) & (purr) & (purr)\\
\bottomrule
\end{tabular}
}
\caption{The examples of mCSQA. The English translations are all machine-translated by DeepL. The translated results sometimes are aggregated into one English word due to ignoring source language-specific subtle meaning differences caused by machine translation. This aggregation has also been observed in X-CSQA, which was created using machine translation of CSQA. Hence, X-CSQA could not evaluate fine-grained, language-specific knowledge for each language, but mCSQA can evaluate it because it is created from scratch for each language.}
\label{tab:example-of-mCSQA}
\end{table*}

\section{Prompts for Creating mCSQA}
\label{sec:detail-of-prompts}
The prompts used for creating mCSQA are presented as follows: English in Table~\ref{tab:template-english}, Japanese in Table~\ref{tab:template-japanese}, Chinese in Table~\ref{tab:template-chinese}, German in Table~\ref{tab:template-german}, Portuguese in Table~\ref{tab:template-portuguese}, Dutch in Table~\ref{tab:template-dutch}, French in Table~\ref{tab:template-french} and Russian in Table~\ref{tab:template-russian}.

In each prompt template, the words within the curly brackets are replaced with data-specific terms\footnote{\url{https://peps.python.org/pep-0498/}} before input to the LM.

Furthermore, as discussed in Section~\ref{sec:polyglot-template}, each template was translated exactly to elicit language-specific knowledge of each language. The translations were carried out using both GPT-3.5 and DeepL to ensure there were no semantic differences, with manual fixing applied as needed. We use the OpenAI API's JSON mode\footnote{\url{https://platform.openai.com/docs/guides/text-generation/json-mode}} has facilitated the retrieval of generation results. 

Our findings as a tip, when inputting structured data such as keywords, doing so in a format similar to a programming code like list type, allows us to obtain results that more following the prompt instructions. This improvement can be attributed to the LM's learning to enhance coding abilities, which is believed to have improved its recognition capabilities.


\begin{table*}[t]
\centering
\small
\begin{tabular}{@{}ll@{}}
\toprule
Steps& Prompt (English)  \\
\midrule
\makecell[tl]{
Create \\question \\sentences
}
& \makecell[tl]{
Please create a multiple-choice question with the following conditions:\\
\\
(a) The only correct answer is ["\{correct\}"].\\
(b) The incorrect answers are ["\{distractor1\}", "\{distractor2\}"].\\
(c) Do not use the words ["\{correct\}", "\{distractor1\}", "\{distractor2\}"] in the question.\\
(d) Avoid using superficial information, such as character count.\\
(e) The question ends with a question mark (?).\\
(f) It should be an objective question that can be sufficiently answered with common sense knowledge alone.\\
(g) The question must be a simple and short sentence consisting of only one sentence.\\
\\
Question:
} \\ 
\midrule
\makecell[tl]{
Refine \\question \\ sentences 
}
& \makecell[tl]{
If the original sentence is semantically and grammatically correct, repeat it;\\ if it is unnatural, please rewrite it into a correct and fluent sentence.\\
\\
\{question\}
} \\ 
\midrule
\makecell[tl]{
Add \\additional \\distractors 
} & \makecell[tl]{
Please only add two plausible and natural choices and save them in \{'additional\_choice':[]\}.\\
\\
\lbrack"\{choice1\}", "\{chioce2\}", "\{choice3\}"\rbrack
}\\
\midrule
\makecell[tl]{
Verify \\Qualities
}& \makecell[tl]{
Please select only one alphabet as the answer from the Answer Choices\\ and save it in the format: \{'answer': selected\_answer\}.\\
\\
Q: \{question\} \\
Answer Choices: (A) \{choice\_a\} (B) \{choice\_b\} (C) \{choice\_c\} (D) \{choice\_d\} (E) \{choice\_e\}
}\\
\bottomrule
\end{tabular}
\caption{The prompt templates used to create the mCSQA in the English version.}
\label{tab:template-english}
\end{table*}

\begin{table*}[t]
\centering
\small
\begin{tabular}{@{}ll@{}}
\toprule
Steps& Prompt (Japanese)  \\
\midrule
\makecell[tl]{
Create \\question \\sentences
}
& \makecell[tl]{
\ja{以下の条件を満たす選択肢付きのクイズ問題を作成してください。}\\
\\
\ja{(a) 正解は["\{correct\}"]のみです。}\\
\ja{(b) 不正解は["\{distractor1\}", "\{distractor2\}"]です。}\\
\ja{(c) 問題文に["\{correct\}", "\{distractor1\}", "\{distractor2\}"]という単語を使わないでください。}\\
\ja{(d) 文字数などの表面的な情報の使用を避けてください。}\\
\ja{(e) 問題文は疑問符（？）で終わります。}\\
\ja{(f) 一般常識だけで十分に答えられる客観的な問題である必要があります。}\\
\ja{(g) 問題文は一文のみから成る単純で短い文でなければなりません。}\\
\\
\ja{問題：}
} \\ 
\midrule
\makecell[tl]{
Refine \\question \\ sentences 
}
& \makecell[tl]{
\ja{元の文が意味的・文法的に正しい場合は繰り返す、}\\
\ja{不自然な場合は正しい流暢な文へ書き換えてください。}\\
\\
\{question\}
} \\ 
\midrule
\makecell[tl]{
Add \\additional \\distractors 
} & \makecell[tl]{
\ja{もっともらしい自然な選択肢を2つだけ追加し、}\\
\ja{それらを\{'additional\_choice':[]\}に保存してください。}\\
\\
\lbrack"\{choice1\}", "\{chioce2\}", "\{choice3\}"\rbrack
}\\
\midrule
\makecell[tl]{
Verify \\Qualities
}& \makecell[tl]{
\ja{Answer Choicesから解答となるアルファベットを１つだけ選び、}\\
\ja{次の形式で保存してください：\{'answer': selected\_answer\}。}\\
\\
Q: \{question\} \\
Answer Choices: (A) \{choice\_a\} (B) \{choice\_b\} (C) \{choice\_c\} (D) \{choice\_d\} (E) \{choice\_e\}
}\\
\bottomrule
\end{tabular}
\caption{The prompt templates used to create the mCSQA in the Japanese version.}
\label{tab:template-japanese}
\end{table*}

\begin{table*}[t]
\centering
\small
\begin{tabular}{@{}ll@{}}
\toprule
Steps& Prompt (Chinese)  \\
\midrule
\makecell[tl]{
Create \\question \\sentences
}
& \makecell[tl]{
\zh{请根据以下条件创建一个多项选择题：}\\
\\
\zh{(a) 唯一正确答案是["\{correct\}"]。}\\
\zh{(b) 错误答案是["\{distractor1\}", "\{distractor2\}"]。}\\
\zh{(c) 问题中不得使用["\{correct\}", "\{distractor1\}", "\{distractor2\}"]这些词。}\\
\zh{(d) 避免使用表面信息，如字符数。}\\
\zh{(e) 问题以问号(?)结束。}\\
\zh{(f) 它应该是一个客观的问题，仅凭常识就能充分回答。}\\
\zh{(g) 问题必须是一个简单且短的句子，仅由一句话组成。}\\
\\
\zh{问题：}
} \\ 
\midrule
\makecell[tl]{
Refine \\question \\ sentences 
}
& \makecell[tl]{
\zh{如果原句在语义和语法上正确，请重复它；如果不自然，请将其改写为正确流畅的句子。}\\
\\
\{question\}
} \\ 
\midrule
\makecell[tl]{
Add \\additional \\distractors 
} & \makecell[tl]{
\zh{请只添加两个合理且自然的选择，并将它们保存在 \{'additional\_choice':[]\} 中。}\\
\\
\lbrack"\{choice1\}", "\{chioce2\}", "\{choice3\}"\rbrack
}\\
\midrule
\makecell[tl]{
Verify \\Qualities
}& \makecell[tl]{
\zh{请从Answer Choices中仅选择一个字母作为答案，并以以下格式保存：\{'answer': selected\_answer\}。}\\
\\
Q: \{question\} \\
Answer Choices: (A) \{choice\_a\} (B) \{choice\_b\} (C) \{choice\_c\} (D) \{choice\_d\} (E) \{choice\_e\}
}\\
\bottomrule
\end{tabular}
\caption{The prompt templates used to create the mCSQA in the Chinese version.}
\label{tab:template-chinese}
\end{table*}

\begin{table*}[t]
\centering
\small
\begin{tabular}{@{}ll@{}}
\toprule
Steps& Prompt (German)  \\
\midrule
\makecell[tl]{
Create \\question \\sentences
}
& \makecell[tl]{
Bitte erstellen Sie eine Multiple-Choice-Frage mit folgenden Bedingungen:\\
\\
(a) Die einzig richtige Antwort ist ["\{correct\}"].\\
(b) Die falschen Antworten sind ["\{distractor1\}", "\{distractor2\}"].\\
(c) Verwenden Sie in der Frage nicht die Wörter ["\{correct\}", "\{distractor1\}", "\{distractor2\}"].\\
(d) Vermeiden Sie oberflächliche Informationen, wie z.B. die Zeichenanzahl.\\
(e) Die Frage endet mit einem Fragezeichen (?).\\
(f) Es sollte eine objektive Frage sein, die allein mit Allgemeinwissen ausreichend beantwortet werden kann.\\
(g) Die Frage muss ein einfacher und kurzer Satz bestehend aus nur einem Satz sein.\\
\\
Frage:
} \\ 
\midrule
\makecell[tl]{
Refine \\question \\ sentences 
}
& \makecell[tl]{
Wenn der Originalsatz semantisch und grammatikalisch korrekt ist, wiederholen Sie ihn; \\wenn er unnatürlich ist, schreiben Sie ihn bitte in einen korrekten und flüssigen Satz um.\\
\\
\{question\}
} \\ 
\midrule
\makecell[tl]{
Add \\additional \\distractors 
} & \makecell[tl]{
Bitte fügen Sie nur zwei plausible und natürliche Optionen hinzu \\und speichern Sie diese in \{'additional\_choice':[]\}.\\
\\
\lbrack"\{choice1\}", "\{chioce2\}", "\{choice3\}"\rbrack
}\\
\midrule
\makecell[tl]{
Verify \\Qualities
}& \makecell[tl]{
Bitte wählen Sie nur einen Buchstaben als Antwort aus den Answer Choices aus \\und speichern Sie ihn im Format: \{'answer': selected\_answer\}.\\
\\
Q: \{question\} \\
Answer Choices: (A) \{choice\_a\} (B) \{choice\_b\} (C) \{choice\_c\} (D) \{choice\_d\} (E) \{choice\_e\}
}\\
\bottomrule
\end{tabular}
\caption{The prompt templates used to create the mCSQA in the German version.}
\label{tab:template-german}
\end{table*}

\begin{table*}[t]
\centering
\small
\begin{tabular}{@{}ll@{}}
\toprule
Steps& Prompt (Portuguese)  \\
\midrule
\makecell[tl]{
Create \\question \\sentences
}
& \makecell[tl]{
Por favor, crie uma pergunta de múltipla escolha com as seguintes condições:\\
\\
(a) A única resposta correta é ["\{correct\}"].\\
(b) As respostas incorretas são ["\{distractor1\}", "\{distractor2\}"].\\
(c) Não use as palavras ["\{correct\}", "\{distractor1\}", "\{distractor2\}"] na pergunta.\\
(d) Evite usar informações superficiais, como a contagem de caracteres.\\
(e) A pergunta termina com um ponto de interrogação (?).\\
(f) Deve ser uma pergunta objetiva que pode ser suficientemente \\respondida apenas com conhecimento de senso comum.\\
(g) A pergunta deve ser uma frase simples e curta, consistindo de apenas uma frase.\\
\\
Pergunta:
} \\
\midrule
\makecell[tl]{
Refine \\question \\ sentences 
}
& \makecell[tl]{
Se a frase original estiver semanticamente e gramaticalmente correta, repita-a; \\se for pouco natural, por favor, reescreva-a em uma frase correta e fluente.\\
\\
\{question\}
} \\ 
\midrule
\makecell[tl]{
Add \\additional \\distractors 
} & \makecell[tl]{
Por favor, adicione apenas duas escolhas plausíveis e naturais e salve-as em \{'additional\_choice':[]\}. \\
\\
\lbrack"\{choice1\}", "\{chioce2\}", "\{choice3\}"\rbrack
}\\
\midrule
\makecell[tl]{
Verify \\Qualities
}& \makecell[tl]{
Por favor, selecione apenas uma letra como resposta das Answer Choices \\e salve no formato:  \{'answer': selected\_answer\}. \\
\\
Q: \{question\} \\
Answer Choices: (A) \{choice\_a\} (B) \{choice\_b\} (C) \{choice\_c\} (D) \{choice\_d\} (E) \{choice\_e\}
}\\
\bottomrule
\end{tabular}
\caption{The prompt templates used to create the mCSQA in the Portuguese version.}
\label{tab:template-portuguese}
\end{table*}

\begin{table*}[t]
\centering
\small
\begin{tabular}{@{}ll@{}}
\toprule
Steps& Prompt (Dutch)  \\
\midrule
\makecell[tl]{
Create \\question \\sentences
}
& \makecell[tl]{
Maak alstublieft een meerkeuzevraag met de volgende voorwaarden:\\
\\
(a) Het enige juiste antwoord is ["\{correct\}"].\\
(b) De onjuiste antwoorden zijn ["\{distractor1\}", "\{distractor2\}"].\\
(c) Gebruik de woorden ["\{correct\}", "\{distractor1\}", "\{distractor2\}"] niet in de vraag.\\
(d) Vermijd het gebruik van oppervlakkige informatie, zoals het aantal tekens.\\
(e) De vraag eindigt met een vraagteken (?).\\
(f) Het moet een objectieve vraag zijn die alleen met algemene kennis voldoende beantwoord kan worden.\\
(g) De vraag moet een eenvoudige en korte zin zijn die uit slechts één zin bestaat.\\
\\
Vraag:
}\\
\midrule
\makecell[tl]{
Refine \\question \\ sentences 
}
& \makecell[tl]{
Als de originele zin semantisch en grammaticaal correct is, herhaal deze dan; \\als het onnatuurlijk is, herschrijf het dan naar een correcte en vloeiende zin.\\
\\
\{question\}
} \\ 
\midrule
\makecell[tl]{
Add \\additional \\distractors 
} & \makecell[tl]{
Voeg alstublieft slechts twee aannemelijke en natuurlijke keuzes toe en sla ze op in \{'additional\_choice':[]\}. \\
\\
\lbrack"\{choice1\}", "\{chioce2\}", "\{choice3\}"\rbrack
}\\
\midrule
\makecell[tl]{
Verify \\Qualities
}& \makecell[tl]{
Selecteer alstublieft slechts één letter als antwoord uit de Answer Choices\\ en sla het op in het formaat: \{'answer': selected\_answer\}.\\
\\
Q: \{question\} \\
Answer Choices: (A) \{choice\_a\} (B) \{choice\_b\} (C) \{choice\_c\} (D) \{choice\_d\} (E) \{choice\_e\}
}\\
\bottomrule
\end{tabular}
\caption{The prompt templates used to create the mCSQA in the Dutch version.}
\label{tab:template-dutch}
\end{table*}

\begin{table*}[t]
\centering
\small
\begin{tabular}{@{}ll@{}}
\toprule
Steps& Prompt (French)  \\
\midrule
\makecell[tl]{
Create \\question \\sentences
}
& \makecell[tl]{
Veuillez créer une question à choix multiples avec les conditions suivantes :\\
\\
(a) La seule bonne réponse est ["\{correct\}"].\\
(b) Les réponses incorrectes sont ["\{distractor1\}", "\{distractor2\}"].\\
(c) Ne pas utiliser les mots ["\{correct\}", "\{distractor1\}", "\{distractor2\}"] dans la question.\\
(d) Évitez d'utiliser des informations superficielles, telles que le nombre de caractères.\\
(e) La question se termine par un point d'interrogation (?).\\
(f) Il doit s'agir d'une question objective qui peut être suffisamment répondue avec le seul sens commun.\\
(g) La question doit être une phrase simple et courte composée d'une seule phrase.\\
\\
Question : \\
}\\
\midrule
\makecell[tl]{
Refine \\question \\ sentences 
}
& \makecell[tl]{
Si la phrase originale est correcte sémantiquement et grammaticalement, répétez-la ;\\ si elle est peu naturelle, veuillez la reformuler en une phrase correcte et fluide. \\
\\
\{question\}
} \\ 
\midrule
\makecell[tl]{
Add \\additional \\distractors 
} & \makecell[tl]{
Veuillez ajouter seulement deux choix plausibles et naturels et les enregistrer dans \{'additional\_choice':[]\}. \\
\\
\lbrack"\{choice1\}", "\{chioce2\}", "\{choice3\}"\rbrack
}\\
\midrule
\makecell[tl]{
Verify \\Qualities
}& \makecell[tl]{
Veuillez sélectionner uniquement une lettre comme réponse parmi les Answer Choices\\ et enregistrez-la dans le format : \{'answer': selected\_answer\}. \\
\\
Q: \{question\} \\
Answer Choices: (A) \{choice\_a\} (B) \{choice\_b\} (C) \{choice\_c\} (D) \{choice\_d\} (E) \{choice\_e\}
}\\
\bottomrule
\end{tabular}
\caption{The prompt templates used to create the mCSQA in the French version.}
\label{tab:template-french}
\end{table*}

\begin{table*}[t]
\centering
\small
\begin{tabular}{@{}ll@{}}
\toprule
Steps& Prompt (Russian)  \\
\midrule
\makecell[tl]{
Create \\question \\sentences
}
& \makecell[tl]{
\textcyr{Пожалуйста, создайте вопрос с несколькими вариантами ответа с учетом следующих условий:}\\
\\
(a) \textcyr{Единственный правильный ответ - }["\{correct\}"]\textcyr{.}\\
(b) \textcyr{Неправильные ответы - }["\{distractor1\}", "\{distractor2\}"]\textcyr{.}\\
(c) \textcyr{Не используйте слова }["\{correct\}", "\{distractor1\}", "\{distractor2\}"]\textcyr{ в вопросе.}\\
(d) \textcyr{Избегайте использования поверхностной информации, такой как количество символов.}\\
(e) \textcyr{Вопрос заканчивается вопросительным знаком (?).}\\
(f) \textcyr{Это должен быть объективный вопрос,}\\ \textcyr{на который можно достаточно ответить только с помощью здравого смысла.}\\
(g) \textcyr{Вопрос должен быть простым и коротким, состоящим только из одного предложения.}\\
\\
\textcyr{Вопрос: }\\
}\\
\midrule
\makecell[tl]{
Refine \\question \\ sentences 
}
& \makecell[tl]{
\textcyr{Если исходное предложение семантически и грамматически правильно, повторите его;}\\ \textcyr{если оно звучит ненатурально, пожалуйста,}\\ \textcyr{перепишите его на корректный и свободно звучащий язык.} \\
\\
\{question\}
} \\ 
\midrule
\makecell[tl]{
Add \\additional \\distractors 
} & \makecell[tl]{
\textcyr{Пожалуйста, добавьте только два правдоподобных и естественных выбора} \\
\textcyr{и сохраните их в} \{'additional\_choice':[]\}.\\
\\
\lbrack"\{choice1\}", "\{chioce2\}", "\{choice3\}"\rbrack
}\\
\midrule
\makecell[tl]{
Verify \\Qualities
}& \makecell[tl]{
\textcyr{Пожалуйста, выберите только одну букву алфавита в качестве ответа из Answer Choices и} \\ \textcyr{сохраните её в формате:} \{'answer': selected\_answer\}. \\
\\
Q: \{question\} \\
Answer Choices: (A) \{choice\_a\} (B) \{choice\_b\} (C) \{choice\_c\} (D) \{choice\_d\} (E) \{choice\_e\}
}\\
\bottomrule
\end{tabular}
\caption{The prompt templates used to create the mCSQA in the Russian version.}
\label{tab:template-russian}
\end{table*}

\end{document}